\documentclass[runningheads]{llncs}

\usepackage{eccv}

\usepackage{eccvabbrv}

\usepackage{graphicx}
\usepackage{booktabs}
\usepackage{physics}
\usepackage{caption}
\usepackage{subcaption}
\usepackage{multirow}
\usepackage{amsmath}
\usepackage{amssymb}
\usepackage[dvipsnames]{xcolor}

\usepackage[accsupp]{axessibility}  % Improves PDF readability for those with disabilities.

\def\METHODNAME{KEEP}

\newcommand{\lichongyi}[1]{\textbf{\color{cyan}(CL: {#1})}}
\newcommand{\emp}[1]{\noindent\textbf{#1}}
\newcommand{\red}[1]{{\color{red}#1}}

\newcommand{\blue}[1]{\textcolor{blue}{#1}} % revised sentence.

\usepackage{hyperref}

\usepackage{orcidlink}

\begin{document}

% ---------------------------------------------------------------
% TODO REVIEW: Replace with your title
\title{Kalman-Inspired Feature Propagation for \\ Video Face Super-Resolution} 

% TODO REVIEW: If the paper title is too long for the running head, you can set
% an abbreviated paper title here. If not, comment out.
\titlerunning{KEEP}

% TODO FINAL: Replace with your author list. 
% Include the authors' OCRID for the camera-ready version, if at all possible.
\author{Ruicheng Feng\orcidlink{0000-0003-4544-3078} \and
Chongyi Li\orcidlink{0000-0003-2609-2460} \and
Chen Change Loy\orcidlink{0000-0001-5345-1591}}

% TODO FINAL: Replace with an abbreviated list of authors.
\authorrunning{R.~Feng et al.}
% First names are abbreviated in the running head.
% If there are more than two authors, 'et al.' is used.

% TODO FINAL: Replace with your institution list.
\institute{S-Lab, Nanyang Technological University, Singapore\\
\email{\{ruicheng002, ccloy\}@ntu.edu.sg}\\
\email{lichongyi25@gmail.com}
}

\maketitle
% \maketitle
% \vspace{-0.9cm}
% \begin{center}
%     \centering
%         \includegraphics[width=.98\linewidth]{figs/teaser.pdf}
%     % \vskip 0.1cm
%     \captionof{figure}{\textbf{Comparing main VFSR strategies.}
%     We show seven frames with an interval of $6$.
%     Generic VSR model BasicVSR~\cite{chan2021basicvsr} fails to reconstruct facial components faithfully.
%     Single-image FSR model CodeFormer~\cite{zhou2022codeformer} hallucinates unnatural and inconsistent face details.
%     Our method, in contrast, enables consistent restoration of low-quality face video while preserving temporal coherence across frames.}
%     \label{fig:teaser}
% \end{center}

\begin{figure}[t]
    \centering
    \includegraphics[width=.98\linewidth]{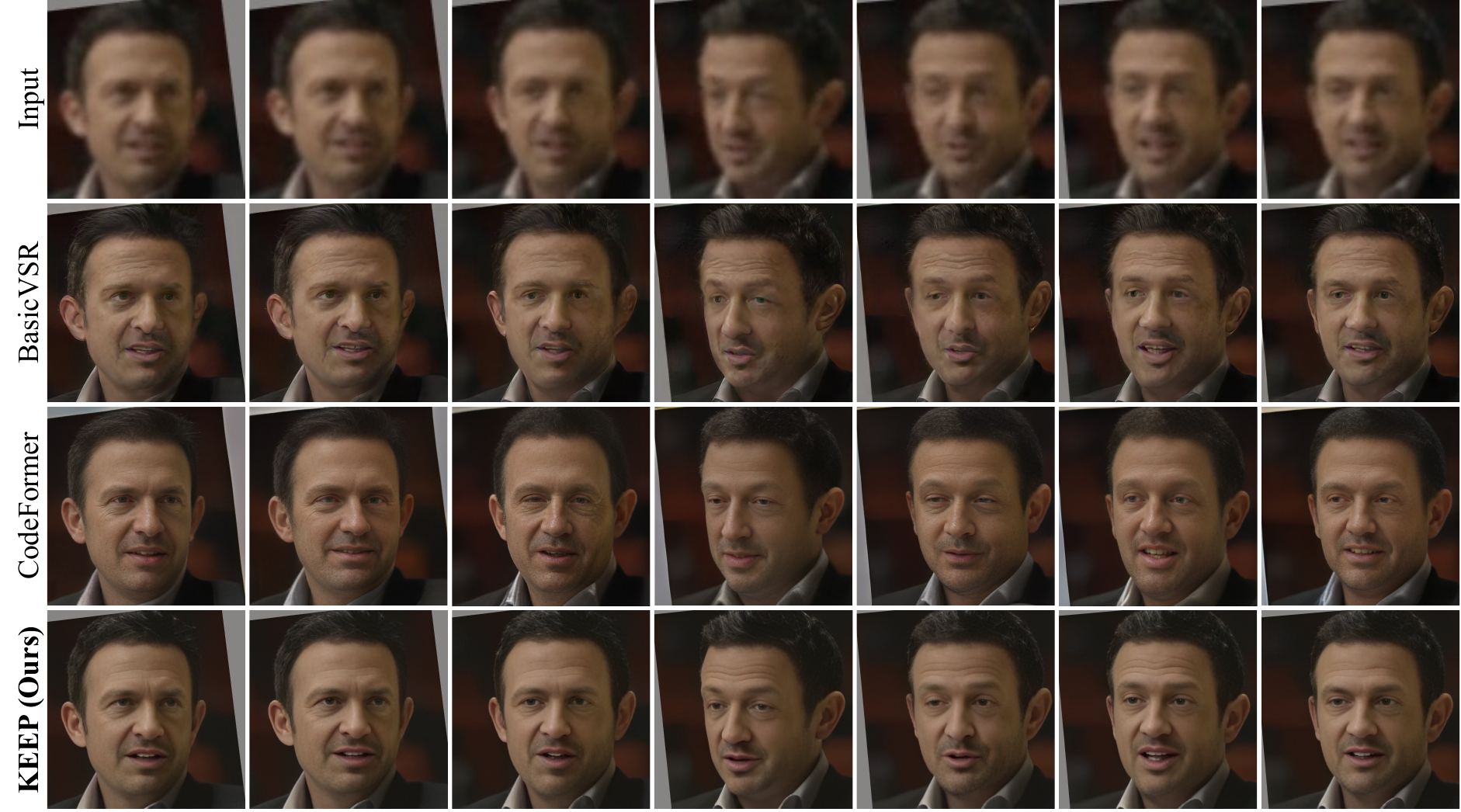}
    % \vspace{-0.2cm}
    \caption{
        \textbf{Comparing main VFSR strategies.}
        We show seven frames with an interval of $6$.
        Generic VSR model BasicVSR~\cite{chan2021basicvsr} fails to reconstruct facial components faithfully.
        Single-image FSR model CodeFormer~\cite{zhou2022codeformer} hallucinates unnatural and inconsistent face details.
        Our method, in contrast, enables consistent restoration of low-quality face video while preserving temporal coherence across frames.
    }
    \label{fig:teaser}
    % \vspace{-0.7cm}
\end{figure}
\begin{abstract}
Despite the promising progress of face image super-resolution,  video face super-resolution remains relatively under-explored. 
Existing approaches either adapt general video super-resolution networks to face datasets or apply established face image super-resolution models independently on individual video frames. 
These paradigms encounter challenges either in reconstructing facial details or maintaining temporal consistency.
To address these issues, we introduce a novel framework called Kalman-inspired Feature Propagation (\METHODNAME), designed to maintain a stable face prior over time.
The Kalman filtering principles offer our method a recurrent ability to use the information from previously restored frames to guide and regulate the restoration process of the current frame.
Extensive experiments demonstrate the effectiveness of our method in capturing facial details consistently across video frames.
Code and video demo are available at \url{https://jnjaby.github.io/projects/KEEP/}.
\end{abstract}
\section{Introduction}
\label{sec:intro}

\if 0
The evolution of face image restoration models \cite{zhou2022codeformer,chan2021glean,wang2021towards,wang2022restoreformer} has demonstrated extraordinary capability in Face super-resolution (FSR), which aims at recovering high-resolution (HR) face images from their severely degraded counterparts. 
Impressive efforts have been devoted to exploiting prior information, \eg, geometric facial priors \cite{chen2021progressive,chen2018fsrnet,yu2018face}, reference priors \cite{li2018learning,li2020blind,li2020enhanced}, generative priors \cite{chan2021glean,chan2022glean,wang2021towards,yang2021gan}, and codebook priors \cite{zhou2022codeformer,gu2022vqfr,wang2022restoreformer} that remarkably push the performance and generate realistic face images.
However, this progress is mostly restricted to images, and expanding it to Video Face Super-Resolution (VFSR) is still underexplored.
Despite the great benefits of video face restoration in practical applications, \eg, video surveillance, old series, and old movies, VFSR remains an area of limited exploration.
\fi

The field of Face Super-Resolution (FSR), which focuses on reconstructing high-resolution (HR) face images from highly degraded versions, has witnessed remarkable progress.
In particular, numerous studies have successfully leveraged various types of prior information, such as geometric facial priors \cite{chen2021progressive,chen2018fsrnet,yu2018face}, reference priors \cite{li2018learning,li2020blind,li2020enhanced}, generative priors \cite{chan2021glean,chan2022glean,wang2021towards,yang2021gan}, and codebook priors \cite{zhou2022codeformer,gu2022vqfr,wang2022restoreformer}. These approaches have significantly advanced the realism and quality of generated face images.
However, the majority of these studies are confined to still images, with the extension to Video Face Super-Resolution (VFSR) remaining relatively under-explored. Despite the substantial potential benefits of video face restoration in various practical domains, such as the restoration of old films, VFSR has yet to receive the same level of attention and development as its image-based counterpart. 

\if 0
There are two intuitive solutions to implement VFSR.
One involves training general Video Super-Resolution (VSR) networks, \eg, EDVR~\cite{wang2019edvr}, BasicVSR~\cite{chan2021basicvsr}, BasicVSR++~\cite{chan2022basicvsr++}, and RVRT~\cite{liang2022recurrent}, on large-scale face video dataset \cite{Nagrani2017VoxCelebAL, xie2022vfhq}. This kind of approach is designed to exploit temporal information and feature propagation across frames.
%which, by design, involves exploiting temporal information and feature propagation across frames.
These methods, however, are not originally designed for face restoration and are unable to recover rich facial details in the absence of expressive priors, particularly in processing severely degraded cases. \lichongyi{we may need to give some justifications in figure 1 or experiments.}
Another solution is to directly employ off-the-shelf face image SR models and generate HR images on a frame-by-frame basis.
There remains a critical challenge: non-temporal image-based SR models will impose temporal inconsistency issues.
One possible explanation is that FSR is highly ill-posed, and priors are insufficient to ensure appearance coherency in the whole video.
Particularly, a corrupted image could correspond to various HR images and thus result in inconsistent structures between the independently generated video frames.
\fi

Two main strategies emerge for implementing VFSR. The first approach involves adapting general Video Super-Resolution (VSR) networks, such as EDVR \cite{wang2019edvr}, BasicVSR~\cite{chan2021basicvsr}, BasicVSR++\cite{chan2022basicvsr++}, and RVRT~\cite{liang2022recurrent}, to large-scale face video datasets \cite{Nagrani2017VoxCelebAL, xie2022vfhq}. These methods exploit temporal information and propagate features across video frames. However, they are not specifically tailored for face restoration and often fall short in reconstructing detailed facial features, particularly in severely degraded scenarios, as depicted in Fig.~\ref{fig:teaser}(second row).
The second method involves applying existing face image SR models to process each video frame independently. This frame-by-frame approach, while straightforward, introduces temporal inconsistencies in the video, as demonstrated in Fig.~\ref{fig:teaser}(third row). This problem arises because FSR is inherently ill-posed and the existing priors may not suffice to maintain appearance consistency throughout the video sequence. Specifically, a single degraded image could correspond to multiple high-resolution interpretations, leading to discrepancies and inconsistent structures across independently processed video frames.

In this paper, we wish to devise an effective framework for maintaining a stable face prior over time for VFSR. We use CodeFormer~\cite{zhou2022codeformer}, a representative model that exploits codebook priors for FSR, to demonstrate how face priors can be consistently preserved across different time frames.
A pivotal aspect of our approach is the idea that frames previously restored can act as references, guiding and regulating the restoration process of the current frame. This strategy helps minimize the divergence between consecutive frames. Moreover, this reliance on previously restored frames naturally suggests a recurrent framework, enabling the effective use of information from past restorations.
This intuition aligns closely with Kalman filtering principles, or linear quadratic estimation, which involves using a sequence of time-based measurements, despite their statistical noise and inaccuracies, to produce more accurate estimates of unknown variables than would be possible with a single measurement. 
Similarly, in VFSR, faces observed over time are often noisy and inaccurate, making them suitable for refinement using Kalman filtering techniques.

\if 0
In addition, the textures and details of each frame are conditioned on the discrete latent codes, which are not specific enough to preserve the consistency of local structures (\eg, wrinkles, hairs) across frames.
Thus, we further adopt cross-frame attention to the generator to enhance local coherency in the regions with rich details.
Our proposed V-CodeFormer strikes a great balance between the quality and consistency of video face restoration.
\ruicheng{Show two more advantages here: 1) our model shows better robustness to severe degradation (limited performance drop on mild, medium, and hard splits dataset) since it captures information from neighboring frames, compared to single-image models. 2) Our model outperforms other methods on non-frontal view faces.}
\fi

Driven by these insights, we formulate a novel method, \textbf{K}alman-inspired f\textbf{E}atur\textbf{E} \textbf{P}ropagation (\METHODNAME), which recurrently updates the current latent state in CodeFormer by incorporating information from preceding frames. This method of temporal propagation within the latent space ensures the stability of the face prior over time, thereby capturing facial details that consistently match in appearance. %Consequently, our approach effectively maintains a uniform appearance across video frames.
%
%\METHODNAME~is further enhanced by strategies designed to preserve local details such as wrinkles and hair across frames. %This is achieved through the use of cross-frame attention in the generator.
%
The effectiveness of \METHODNAME~is shown in Fig.~\ref{fig:teaser}, where it is evident that our method delivers high-quality restoration with superior consistency, compared to both generic video restoration methods fine-tuned for faces and approaches that restore frames independently. Please refer to the supplementary video to appreciate the superiority of our approach in terms of temporal consistency.
A key advantage of \METHODNAME~is its robustness in handling severe video-based degradation, outperforming single-image models. In addition, our model demonstrates enhanced performance on non-frontal faces by providing more stable estimations of face priors.

\if 0
The main contributions of this work are as follows:
\begin{itemize}
    \item We repurpose image face super-resolution model to video FSR, named \textit{V-CodeFormer} which incorporates Kalman filter network to utilize facial prior information while preserving temporal consistency.
    \item We adopt cross-frame attention \lichongyi{cross-frame attention is not now. try to rephrase the second contribution?} to maintain local temporal consistency and recover coherent details.
    \item Experimental results demonstrate that our proposed approach outperforms existing methods on challenge tasks. \lichongyi{this contribution is not strong.}
\end{itemize}
\fi

In summary, the main contribution of this work is a novel framework for maintaining a stable and meaningful face prior for VFSR. While we demonstrate its application using the CodeFormer method as a case study, the underlying principles of our framework, inspired by the Kalman filtering approach, are applicable to other approaches.
Extensive experimental results on the VFHQ dataset~\cite{xie2022vfhq} and real-world data demonstrate the effectiveness of our approach in improving both the fidelity and coherence of VFSR outputs. 
Compared to other state-of-the-art methods, our model achieves superior performance with a large margin of $0.8$ dB in PSNR, while also significantly maintaining temporal coherence.

\section{Related Work}
\label{sec:related_work}

\noindent
\textbf{Blind Face Restoration.}
Blind face restoration aims at recovering severely degraded face images in the wild.
Unlike natural images, faces are highly structured. 
This property allows researchers to incorporate prior information into the restoration models, which have demonstrated remarkable progress in capabilities to restore high-quality faces.
Most existing FSR methods can be categorized into four classes: geometric priors, reference priors, generative priors, and codebook priors.
\textit{Geometric priors} usually include facial elements, such as face landmarks \cite{chen2018fsrnet}, parsing maps \cite{chen2021progressive}, and facial component heatmaps \cite{yu2018face}. %Yet, these priors are unreliable on corrupted face images since most estimation algorithms are designed for high-quality images.
Another major line is \textit{reference-based methods} that require high-quality exemplar images. GFRNet \cite{li2018learning} and ASFFNet \cite{li2020enhanced} leverage a warped high-quality image to extract rich details to improve facial detail restoration. 
DFDNet \cite{li2020blind} constructs deep dictionaries with facial components from large-scale images to recover fine details.
\textit{Generative priors} from pre-trained GAN, \eg, StyleGAN~\cite{karras2019style} and StyleGAN2~\cite{karras2020analyzing}, are employed through iterative latent optimization of GAN inversion \cite{gu2020image,menon2020pulse,pan2021exploiting}. Still, they produce face images with low fidelity and are computationally expensive.
To address this, GLEAN \cite{chan2021glean}, GPEN~\cite{yang2021gan}, and GFPGAN~\cite{wang2021towards} integrate generative priors into encoder-decoder architectures, which estimate latent priors in one-forward pass. These methods achieve great trade-off between quality and fidelity but usually fail when the corruption is severe. 
\textit{Codebook priors} \cite{zhou2022codeformer,gu2022vqfr,wang2022restoreformer} can be regarded as a special case of generative priors. In contrast to continuous generative priors, they squeeze the latent space into a small finite codebook space and improve the robustness to severe degradation.
However, most existing FSR methods are image-based and thus they cannot guarantee temporally consistent details for VFSR.

\noindent
\textbf{Video Super-Resolution.}
% In VSR, it is essential to facilitate frame recovery with other frames in the sequence. Consequently
Most existing video restoration techniques can be categorized into two paradigms based on their parallelizability: parallel and recurrent methods.
\textit{Parallel models} estimate all frames simultaneously, and the restoration of each frame does not rely on the update of other frames.
These methods typically involve feature extraction, feature alignment, feature fusion, and reconstruction.
%They primarily employ 2D or 3D CNN for feature extraction.
%For feature alignment, concurrent approaches either use explicit (optical flow estimation) or implicit (deformable convolutions) alignment modules to aggregate information from neighboring frames.
Representative works, FSTRN \cite{li2019fast} and VESCPN \cite{caballero2017real}, introduce fast spatio-temporal networks using 3D convolutions to enhance alignment, combining motion compensation and super-resolution.
EDVR \cite{wang2019edvr} and TDAN \cite{tian2020tdan} leverage deformable convolutions for aligning adjacent frames.
%However, aligning all frames toward the center frame results in quadratic complexity and poses challenges for extending to long-sequence videos.
Besides, Transformer-based methods \cite{liu2022video,liang2024vrt} are proposed to reconstruct all frames simultaneously by jointly extracting, aligning, and fusing features.
In addition, RVRT \cite{liang2022recurrent} and TTVSR \cite{liu2022learning} integrate optical flow into Transformer and enable long-range models in videos.
Empowered by the great expressive capability of Transformer, this line of work exhibits remarkable performance improvements over previous methods. However, they suffer from large model sizes and high memory consumption.
\textit{Recurrent methods} \cite{chan2021basicvsr,huang2015bidirectional,isobe2020revisiting} do not aggregate information solely from adjacent frames.
Instead, they maintain hidden states to convey relevant information from previous frames and propagate latent features sequentially, accumulating information for later restoration.
For example, RBPN \cite{haris2019recurrent} treated each frame as a separate source, combined iteratively in a refinement framework.
RSDN \cite{isobe2020video} divided the input into structure and detail components, proposing a two-stream structure-detail block to learn textures.
BasicVSR \cite{chan2021basicvsr} and BasicVSR++ \cite{chan2022basicvsr++} fused bidirectional hidden states from both past and future frames, bring significant improvements.
They aim to fully utilize information from the entire sequence, synchronously updating the hidden state through the weights of the reconstruction network.
Due to the recurrent nature of feature propagation, recurrent methods experience information loss. % and are not easily deployable in a distributed manner.

\noindent
\textbf{Video Consistency.}
In addition to VSR, previous works also attempt to inflate image models into video models \cite{tzaban2022stitch,wu2023tune,yang2023rerender,xu2022temporally} or improve temporal consistency with the implicit representation of the given videos \cite{lei2020blind, ouyang2024codef,lei2023blind}.
For example, Stitch-it-in-Time \cite{tzaban2022stitch} discovers locally consistent pivots in the latent space to provide spatially consistent transitions.
Tune-A-Video \cite{wu2023tune} adopts cross-frame attention and fine-tune it on a single video.
All-in-One deflicker \cite{lei2023blind} learns a neural atlas for each video to solve long-term inconsistency.
These works either require post-processing or optimization on a video basis.
In contrast, we aim at repurposing image FSR models and enforcing temporal consistency in the restored face videos.
\section{Methodology}
In a corrupted face video, local textures and facial details are irrevocably lost.
Therefore, the latent codes, usually estimated by an encoder, are inaccurate to match the real underlying priors of ground-truth video.
Different from image-based autoencoder, models in the video domain enable us to exploit evidence accumulated from preceeding frames for better restoration.
In this paper, we repurpose the image-based CodeFormer for video FSR and propose \METHODNAME~to estimate stable face priors in latent space over time given the noisy and inaccurate estimations, which consequently enable temporally coherent restoration.
An appealing idea to realize \METHODNAME~is through reformulating the method in a Kalman Filter framework.

\subsection{Formulation}
\paragraph{\textbf{State Space Model.}}
We consider observations of low-quality (LQ) video sequence $X=\{\vb*{x}_t\}^T_{t=1}$ with length of $T$, where $\vb*{x}_t\in\mathbb{R}^{H\times W\times 3}$, and underlying high-quality (HQ) sequences $Y=\{\vb*{y}_t\}^T_{t=1}$.
Kalman filter \cite{kalman1960new} assumes linear dynamic systems that are characterized by a state space model driven by Gaussian noise
\begin{equation}
    \vb*{y}_t = \vb*{F}_t \vb*{y}_{t-1} + \vb*{q}_t,
\end{equation}
where $\vb*{F}_t$ is the transition matrix and $\vb*{q}_t$ denotes process noise drawn from Gaussian noise.
The observation $\vb*{x}_t$ is measured by 
\begin{equation}
    \vb*{x}_t = \vb*{H} \vb*{y}_t+\vb*{r}_t,
\end{equation}
where $\vb*{H}$ is the measurement matrix and $\vb*{r}_t$ represents measurement noise.
However, the linear assumption does not hold in some complex real-world scenarios.
Hence, the non-linear Kalman filter can be reformulated as 
\begin{align}
    & \vb*{y}_t = d(\vb*{y}_{t-1}, \vb*{q}_t)\\
    & \vb*{x}_t = h(\vb*{y}_t) + \vb*{r}_t,
\end{align}
where $d(\cdot)$ and $h(\cdot)$ are non-linear transition and measurement models.
%In this problem, 
Specifically, $d(\cdot)$ can be represented by any explicit motion estimation (\eg, optical flow), which defines how the current frame transits to the next one.
%In addition, 
$h(\cdot)$ commonly models the degradation in video restoration problems. As opposed to the classical assumptions in Kalman filter, the measurement function $h(\cdot)$ is unknown in a blind setting. This is regarded as \textit{partially} known dynamic models \cite{revach2022kalmannet}.

\begin{figure*}[t]
    \centering
    \includegraphics[width=.98\linewidth]{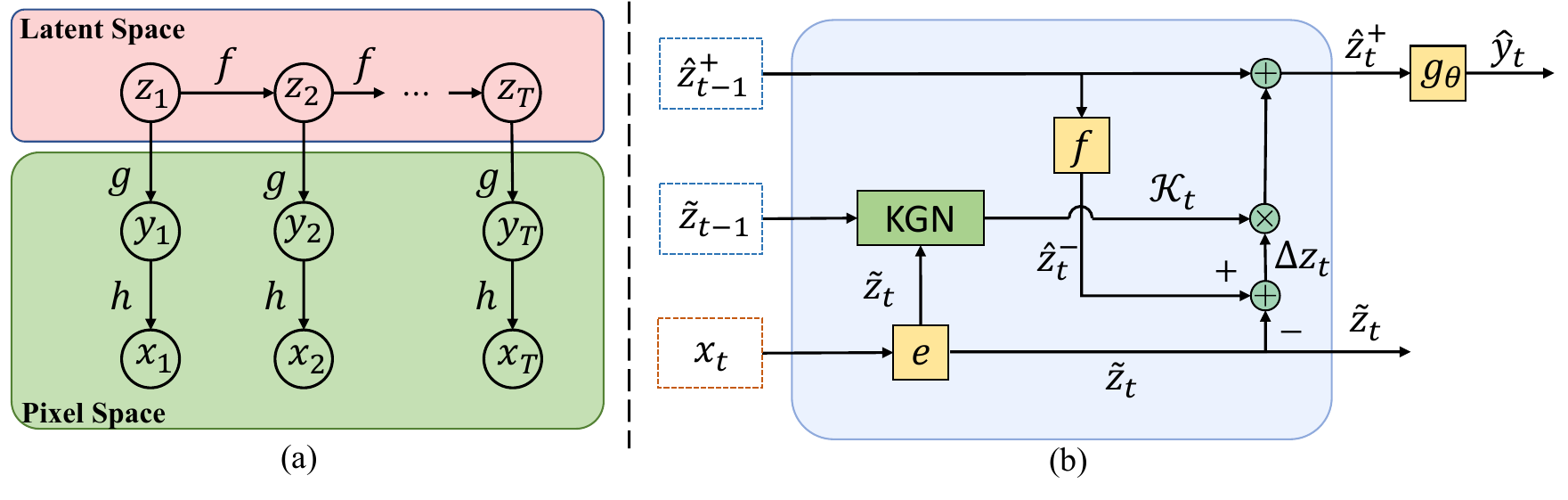}
    % \vspace{-0.2cm}
    \caption{\textbf{(a) Graphical model of state space.} It defines the underlying dynamic system model, where $f$ describes how the latent states $z_t$ transit over time, $g$ is a generative model, and $h$ models the degradation from clean frame $y_t$ to degraded frame $x_t$.
    \textbf{(b) Block diagram of Kalman filter model.}
    In each time step, a predictive state from previous frame $\hat{z}^+_{t-1}$ (Blue dash box) and new observed state of current frame $x_t$ (Red dash box) are fused by Kalman gain $\mathcal{K}_t$ from Kalman Gain Network (KGN) to produce more accurate estimates. The combined state $\hat{z}^+_t$ is then used to generate the estimated clean frame $\hat{y}_t$ by $g_{\theta}$. Note that $\Tilde{z}_{1}$ goes along with $\Tilde{z}_{t-1}$ as an anchor and it is omitted in the diagram for simplicity.
    }
    \label{fig:diagram}
    % \vspace{-.3cm}
\end{figure*}

Inspired by VQGAN~\cite{esser2021taming} and Stable Diffusion~\cite{rombach2022high}, we estimate underlying latent representations $Z=\{\vb*{z}_t\}^T_{t=1}$ such that
$\vb*{z}_t$ can correspond to $\vb*{y}_t$ by a generative model $g_\theta$, given by
\begin{equation}
    \vb*{y}_t = g_\theta(\vb*{z}_t).
\end{equation}
Instead of directly estimating individual pixels, modeling the low-dimensional latent code is computationally more efficient and focuses on more perceptually significant variations of the data. The graphical model is depicted in Fig. \ref{fig:diagram} (a).

\paragraph{\textbf{Kalman Filter Model.}}
The principles of Kalman filter can be formulated by a two-step procedure, \ie, \textit{state prediction} and \textit{state update}.
The overview diagram is illustrated in Fig. \ref{fig:diagram} (b).
In this problem, the observation is a face image $\vb*{x}_t$, and the state is $\vb*{z}_t$.

In the \textit{state prediction} step, the model predicts the prior estimation $\hat{\vb*{z}}_t^-$ of the current state $\vb*{z}_t$ based on the posterior estimation $\hat{\vb*{z}}_{t-1}^+$ of the previous state and the dynamic model.
Specifically, the prior estimation of latent state $\vb*{z}_t$ and estimation of observation $\vb*{x}_t$ are computed as
\begin{align}
    & \hat{\vb*{z}}_t^- = f(\hat{\vb*{z}}_{t-1}^+), \\
    & \hat{\vb*{x}}_t^- = h(g_{\theta}(\hat{\vb*{z}}_t^-)).
\end{align}
The system dynamics $f$ define how the latent state $\vb*{Z}$ evolves over time, and it incorporates any control inputs that might affect the current state.

In the \textit{state update} step, a posterior state estimation $\hat{\vb*{z}}_t^+$ is computed based on the prior estimation and new observations $\vb*{x}_t$ as
\begin{equation}
    \hat{\vb*{z}}_t^+ = \hat{\vb*{z}}_t^- + \mathcal{K}_t \Delta \vb*{z}_t,
    \label{eqn:update}
\end{equation}
where $\mathcal{K}_t$ is conceptually referred to as the \textit{Kalman gain} and $\Delta \vb*{z}_t$ as the \textit{innovation}, \ie, the residual between the prior estimation $\hat{\vb*{z}}_t^-$ and approximation of current state from $\vb*{x}_t$, given by
\begin{equation}
    \Delta\vb*{z}_t = \hat{\vb*{z}}_t^- - \Tilde{\vb*{z}}_t,
\end{equation}
where $\Tilde{\vb*{z}}_t=e(\vb*{x}_t)$.
Note that this formula differs from the original Kalman filter which minimizes the innovation of $\Delta \vb*{x}_t$ (\ie, residual between $\vb*{x}_t$ and $\hat{\vb*{x}}_t^-$), since the assumption of available measurement function $h(\cdot)$ is no longer valid in our setting.
An original measurement system models how observation $\vb*{x}_t$ is derived from the latent state $\vb*{z}_t$, formally $\vb*{x}_t=h(\vb*{z}_t)$.
Inspired by KFNet \cite{zhou2020kfnet}, we directly estimate the state $\vb*{z}_t$ given new observation $\vb*{x}_t$, \ie, mapping $\vb*{x}_t$ to $\Tilde{\vb*{z}}_t$ with an estimator $e$.

The remaining problem is to compute the Kalman gain $\mathcal{K}_t$.
As discussed in KalmanNet \cite{revach2022kalmannet}, covariance estimation is intractable when dealing with high-dimensional signals.
Additionally, the second-order statistical moments are only used for calculating Kalman gain.
Inspired by this, we directly learn the gains from the data distribution and do not explicitly maintain an estimation of covariances.
Additionally, we follow \cite{yang2023rerender} to include $\Tilde{\vb*{z}}_1$ of the first frame as anchor into KGN for Kalman gain estimation.
Then, the final predicted $\hat{\vb*{y}}_t$ can be derived by 
\begin{equation}
    \hat{\vb*{y}}_t = g_\theta(\hat{\vb*{z}}_t^+).
\end{equation}

\subsection{Parameterized Models}
\begin{figure*}[t]
    \centering
    % \draftfig{0.34}{0.94}{V-CodeFormer}
    \includegraphics[width=.9\linewidth]{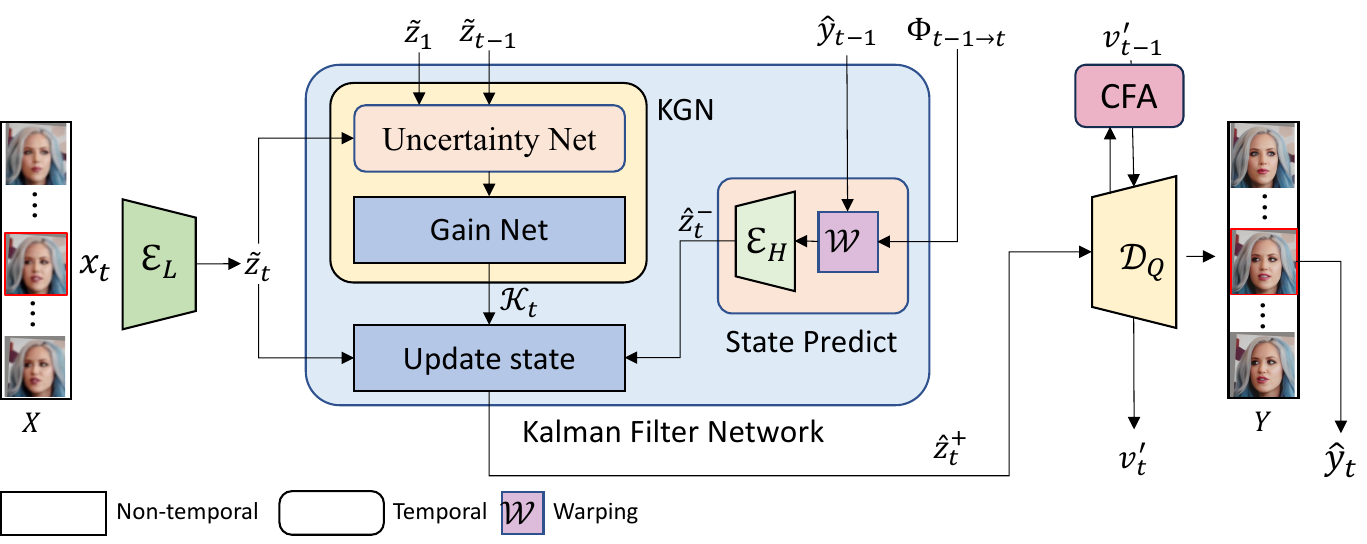}
    % \vspace{-0.2cm}
    \caption{\textbf{Overview of the proposed \METHODNAME.} 
        It consists of four modules: encoder $\mathcal{E}_L$, decoder $\mathcal{D}_Q$, Kalman filter network, and CFA.
        We illustrate the information flow in one timestep.}
    \label{fig:architecture}
    % \vspace{-.3cm}
\end{figure*}
As shown in Fig. \ref{fig:diagram} (b), we will parameterize or define the system dynamics $f$, observation estimator $e$, Kalman Gain Nets (KGN), and generative model $g_\theta$ in this section.
The overall framework is shown in Fig.~\ref{fig:architecture}.

\paragraph{\textbf{Generative Model.}}
The generative model $g_\theta$ is generally parameterized by a backbone of CodeFormer~\cite{zhou2022codeformer}, which consists of a LQ encoder $\mathcal{E}_L$, a HQ encoder $\mathcal{E}_H$, a codebook lookup Transformer and quantization layer $T_Q$, and a decoder $\mathcal{D}$.
For simplicity, $\mathcal{D}_Q$ denotes the decoder with the codebook lookup Transformer and quantization layer absorbed.
Basically, the predicted $\hat{\vb*{y}}_t$ is given by $\mathcal{D}_Q(\hat{\vb*{z}}_t^+)$, and the observed state $\Tilde{\vb*{z}}_t$ is approximated by
\begin{equation}
    \Tilde{\vb*{z}}_t = e(\vb*{x}_t) = \mathcal{E}_L(\vb*{x}_t).
\end{equation}

% The intuition behind this is that the property of VAE will map LQ image $x_t$ and HQ image $y_t$ into the same latent space, \ie, $\mathcal{E}_L(x_t) \approx \mathcal{E}_H(y_t)$.
% Hence, $\mathcal{E}_L(x_t) \approx \mathcal{E}_H(\hat{y}_t)$ will also be true when $\hat{y}_t$ well approximates the underlying $y_t$.

\paragraph{\textbf{State Dynamic System.}}
% In the state prediction step, the filter predicts the prior estimation of the current state $\vb*{z}_t$ of the system (the features of the current frame to be processed) based on the posterior estimation of previous state $\hat{\vb*{z}}_{t-1}^+$ and the system's dynamic model.
The system dynamics define how the system evolves over time, and it incorporates any control inputs that might affect the current state.
The prediction for the state $\vb*{z}_t$ at the current timestep is achieved via state extrapolation.
In particular, given the posterior estimation of previous state $\hat{\vb*{z}}_{t-1}^+$, we define the dynamic model by
\begin{equation}
    \hat{\vb*{z}}_t^- = f(\hat{\vb*{z}}_{t-1}^+) = \mathcal{E}_H(\omega(\mathcal{D}_Q(\hat{\vb*{z}}_{t-1}^+), \Phi_{t-1\rightarrow t})).
\end{equation}
where $\Phi_{t-1\rightarrow t}$ denotes the flow estimated from LQ frames $\vb*{x}_{t-1}$ to $\vb*{x}_t$ and $\omega$ the spatial warping modules.
Specifically, we first decode the estimated code $\hat{\vb*{z}}_{t-1}^+$ of the previous frame to get the estimation of $\hat{\vb*{y}}_{t-1}=\mathcal{D}_Q(\hat{\vb*{z}}_{t-1}^+)$ and warp it to the current frame. Then, it was encoded back in the latent space to obtain the prediction of the current state $\hat{\vb*{z}}_t^-$.
% The intuition behind this is that the property of VAE will map LQ image $x_t$ and HQ image $y_t$ into the same latent space, \ie, $\mathcal{E}_L(x_t) \approx \mathcal{E}_H(y_t)$.
% Hence, $\mathcal{E}_L(x_t) \approx \mathcal{E}_H(\hat{y}_t)$ will also be true when $\hat{y}_t$ well approximates the underlying $y_t$.

\paragraph{\textbf{Kalman Filter System.}}
% Given LQ video sequence $X$, the Kalman filter takes in a sequence of noisy measurements $X$ over time.
% Instead of directly estimating $Y$ in pixel space, we opt to estimate the codes $Z=\{\vb*{z}_t\}^T_{t=1}$ in latent space, since $Y$ are related to the system's state $Z$.
% Then, the estimated $\Tilde{\vb*{y}}_t$ of each frame can be represented by $\Tilde{\vb*{y}}_t=\mathcal{D}(\vb*{z}_t^q)$, where $\vb*{z}_t^q$ is the quantized latent code.
% This formulation allows us to integrate Kalman filtering into CodeFormer.

% Similar to \cite{zhou2022codeformer}, a quantization Transformer $T_q$ is adopted for accurate prediction of discrete codes.
% Hence, Nearest-Neighboring matching is replaced by the Transformer to quantize the code, giving $\vb*{z}_t^q=T_q(\vb*{z}_t)$.
% Moreover, an encoder $\mathcal{E}_L$ for images in the low-quality domain is used to obtain LQ features and shares the same latent space as $\mathcal{E}_H$.
Given the approximated observed state $\Tilde{\vb*{z}}_t$ and prior estimation $\hat{\vb*{z}}_t^-$ from system dynamics, the filter system aims to promote temporal information propagation and maintain stable latent code priors.
In particular, the filter recursively fuses both the estimation to form a more accurate posterior estimate of the current state $\hat{z}_t^+$, which is also known as \textit{state update}.

According to Eqn. \ref{eqn:update}, a more intuitive way to express the updated state is a linear interpolation $\hat{\vb*{z}}_t^+ = \mathcal{K}_t \hat{\vb*{z}}_t^- +  (1 - \mathcal{K}_t)\Tilde{\vb*{z}}_t$ for $\mathcal{K}_t$ normalized in the range of $[0,1]$.
The Kalman Gain $\mathcal{K}_t$ measures the estimated accuracy of the predicted states compared to the approximated observed states, to update the state and reduce the uncertainty.
As illustrated in Fig. \ref{fig:architecture}, the Kalman gain is approximated via Kalman Gain Network (KGN), consisting of two distinct parameterized modules, \ie, uncertainty network and gain network.
The uncertainty associated with the current prediction is implicitly estimated by an uncertainty network constructed by spatial-temporal attention~\cite{wu2023tune} and temporal attention layers (or any other recurrent networks).
Then, a gain network calculates Kalman gain $\mathcal{K}_t$ for each code token.
Please refer to the supplementary files for the detailed architectures.

\subsection{Local Temporal Consistency}
Inspired by \cite{yang2023rerender}, we adopt cross-frame attention (CFA) layers in the decoder to further promote local temporal consistency to regularize the information propagation.
Specifically, given the latent features from the previous frame $v_{t-1}$ and current frame $v_t$. They are projected onto the embedding space and output the features $v_i'$ by
$v_i'=\text{Attn}(Q, K, V)=\text{softmax}(\frac{QK^T}{\sqrt{d}})\cdot V$, where
\begin{equation}
    Q=W_Q\cdot v_t, K=W_K\cdot v_{t-1}, V=W_V\cdot v_{t-1}.
\end{equation}
Intuitively, cross-frame attention modules can be regarded as searching and matching similar patches from the previous frame and fusing them correspondingly.
This module facilitates temporal information propagation in the decoder.
We adopt CFA modules on features of small scale $16$ and $32$ to avoid introducing blur to the decoded results.

% \textbf{Training Objectives.}
% To train the cross-frame attention (CFA) modules and controllable feature Transformation (CFT) in this stage,
% we only relax these two modules and use image-level reconstruction loss $\mathcal{L}_1$, $\mathcal{L}_{per}$ and $\mathcal{L}_{adv}$, which are the same as objectives in Stage I.
% The complete loss for this stage is the sum of above losses weighted with their original weight factors.
% This can further improve the local coherence of our model.

% \subsection{Relations to other Kalman filter networks}
% As discussed in KalmanNet \cite{revach2022kalmannet}, covariance estimation is intractable when dealing with high-dimensional signals. Inspired by this, we bypass this issue by learning the gains from the data distribution.
% With another critical difference being that the measurement matrix is unknown in our blind setting, we reformulate KEEP to fit the problem setting without aligning with the original Kalman filter forcefully.
% Following KFNet \cite{zhou2020kfnet}, we propose a measurement system that directly estimates the state $z_t$ given new observation $x_t$.
\section{Experiments}

\emp{Dataset.}
VFHQ~\cite{xie2022vfhq} contains over $15, 000$ high-quality video clips of diverse interviews and talk shows, where $15,381$ clips are used for training and $50$ clips are reserved for testing. Each sequence consists of $100$ to $900$ frames of resolution $512\times 512$. Following common practice \cite{wang2021towards,xie2022vfhq,chan2022investigating,pan2021deep}, we adopt blind settings in all experiments. Specifically, we apply random blur, resize, and noise as image-based degradations. Moreover, video compression is adopted to control the video quality by changing streaming bitrate.
For a comprehensive evaluation, we synthesize three splits of the VFHQ-Test dataset containing different levels of degradation, denoted as VFHQ-mild, VFHQ-medium, and VFHQ-heavy. They follow the same degradation model but differ in the degree of noise, blur, and compression.
In addition to synthetic data, we also collect real corrupted video for testing. Please refer to the supplementary files for more details.

\emp{Alignment.}
Pre-trained image models for face restoration are trained on the FFHQ dataset \cite{karras2019style}, where each image is automatically cropped and aligned.
Hence, employing pre-trained models requires a similar alignment phase on VFHQ dataset~\cite{xie2022vfhq}.
However, the discrete step (\ie, cropping) is sensitive to the detected locations of landmarks, which can consequently result in unintentional temporal inconsistencies.
Inspired by the work of Fox~\etal~\cite{fox2021stylevideogan} and Tzaban~\etal~\cite{tzaban2022stitch}, we employ a Gaussian lowpass filter over the landmarks.
We find that this smoothing can significantly attenuate the inconsistencies induced by the alignment step. See supplementary files for more details.

\emp{Implementations.}
For all stages of training, we initialize all networks with Kaiming Normal \cite{he2015delving} and train them using Adam optimizer \cite{kingma2014adam},
%with $\beta_1=0.9$, $\beta_2=0.999$, $\theta=10^{-8}$, 
and a batch size of $4$ for all the experiments.
The learning rate is set to $2\times 10^{-4}$ for stages I and II, and $1\times 10^{-4}$ for stage III.
The models are trained with $800k$, $400k$, and $50k$ iterations for three stages, respectively.
We implement our models with PyTorch \cite{paszke2017automatic} and train them using NVIDIA Tesla V100 GPUs.
Hyper-parameters $\lambda_1$, $\lambda_{VGG}$, and $\lambda_{GAN}$ are set to $10^{-2}$, $1$, and $1\times 10^{-2}$.
We use GMFlow \cite{xu2022gmflow} for optical flow estimation.

\emp{Metrics.}
For quantitative evaluation, we evaluate the fidelity of restoration using PSNR, SSIM, and LPIPS~\cite{zhang2018unreasonable}.
In addition, we evaluate the identity preservation scores, termed as \textit{IDS}, by cosine similarity of the off-the-shelf identity detection network ArcFace~\cite{deng2019arcface}.
Besides, we follow Tzaban \etal~\cite{tzaban2022stitch} to measure the pose consistency using Average Keypoint Distance (AKD), which is quantified by the average distance of detected landmarks between the generated and ground-truth video frames.
In addition to the above single-frame quality evaluation, we also measure the fluctuation of identity/landmarks across frames. Hence, $\sigma_{IDS}$ measures the standard deviation of identity similarity over the entire video.
We expect a considerable identity drift in the generated videos without local identity jitter, where $\sigma_{IDS}$ is supposed to be low.
Similarly, we use $\sigma_{AKD}$ to measure the standard deviation of keypoint distances over the video, which quantifies the temporal consistency of the pose.

% evaluate the video’s consistency at the local level.
% identity similarity between pairs of adjacent video frames.
% To account for the effect of inconsistencies in the identity network itself, we normalize these identity preservation scores by the similarity score of each pair of frames
% in the original video. Finally, we average the normalized
% scores over the entire video, and then once again over a set
% of videos. Higher TL-ID scores indicate that the method
% produces smooth results, without considerable local identity jitter.

\subsection{Comparison with State-of-the-Art Methods}
\emp{Baselines.}
We compare our method to two categories of approaches.
i) Image-based Face SR models (CodeFormer~\cite{zhou2022codeformer}, GPEN~\cite{yang2021gan}, GFPGAN~\cite{wang2021towards}, RestoreFormer~\cite{wang2022restoreformer}) are used to generate face videos frame-by-frame.
ii) We retrain the general VSR models (EDVR~\cite{wang2019edvr}, BasicVSR~\cite{chan2021basicvsr}, BasicVSR++~\cite{chan2022basicvsr++}) on VFHQ dataset~\cite{xie2022vfhq}. The degradation settings remain unchanged as our experiments while other training settings follow their original papers.

\begin{table*}[t]
    \centering
    \caption{\textbf{Quantitative comparison on the VFHQ-mild.} \red{Red} and \blue{Blue} indicate the best and the second best results. Full results on other test partitions (medium and heavy) are presented in the supplementary material.}
    \label{tab:sota_comparison}
    % \vspace{-0.1cm}
  \resizebox{.9\linewidth}{!}{
    \begin{tabular}{l||ccc|cc|cc}
    \toprule
        Method & PSNR$\uparrow$ & SSIM$\uparrow$ & LPIPS$\downarrow$ & IDS$\uparrow$ & AKD$\downarrow$ & $\sigma_{IDS}(\times 10^{-2})\downarrow$ & $\sigma_{AKD}\downarrow$ \\
    \midrule
        GPEN \cite{yang2021gan} & 25.5193 & 0.7517 & 0.2988 & 0.7142 & 11.4691 & 4.7416 & 3.5109 \\
        GFPGAN \cite{wang2021towards} & 26.2933 & 0.7795 & 0.2482 & 0.7437 & 10.5467 & 4.5700 & 3.6482 \\
        RestoreFormer \cite{wang2022restoreformer} & 25.5720 & 0.7344 & 0.3195 & 0.7530 & \blue{10.5354} & \blue{4.7159} & \blue{3.4122} \\
        CodeFormer \cite{zhou2022codeformer} & 24.6597 & 0.7454 & 0.2742 & 0.6272 & 11.4983 & 6.3726 & 3.6927 \\
    \midrule
        EDVR \cite{wang2019edvr} & 26.6051 & 0.7858 & 0.2484 & 0.7195 & 11.6220 & 4.8048 & 3.5829 \\
        BasicVSR \cite{chan2021basicvsr} & 26.0458 & 0.7765 & 0.2496 & 0.6973 & 11.3679 & 5.0343 & 3.6054 \\
        BasicVSR++ \cite{chan2022basicvsr++} & \blue{27.1996} & \blue{0.8057} & \blue{0.1958} & \blue{0.7641} & 11.3136 & 5.2543 & 4.6425 \\
        \textbf{KEEP (Ours)} & \red{27.9994} & \red{0.8267} & \red{0.1619} & \red{0.7960} & \red{8.8182} & \red{3.6866} & \red{3.2538} \\
    \bottomrule
    \end{tabular}
    }
    % \vspace{-0.4cm}
\end{table*}
\begin{figure*}[t]
    \centering
    \includegraphics[width=.9\linewidth]{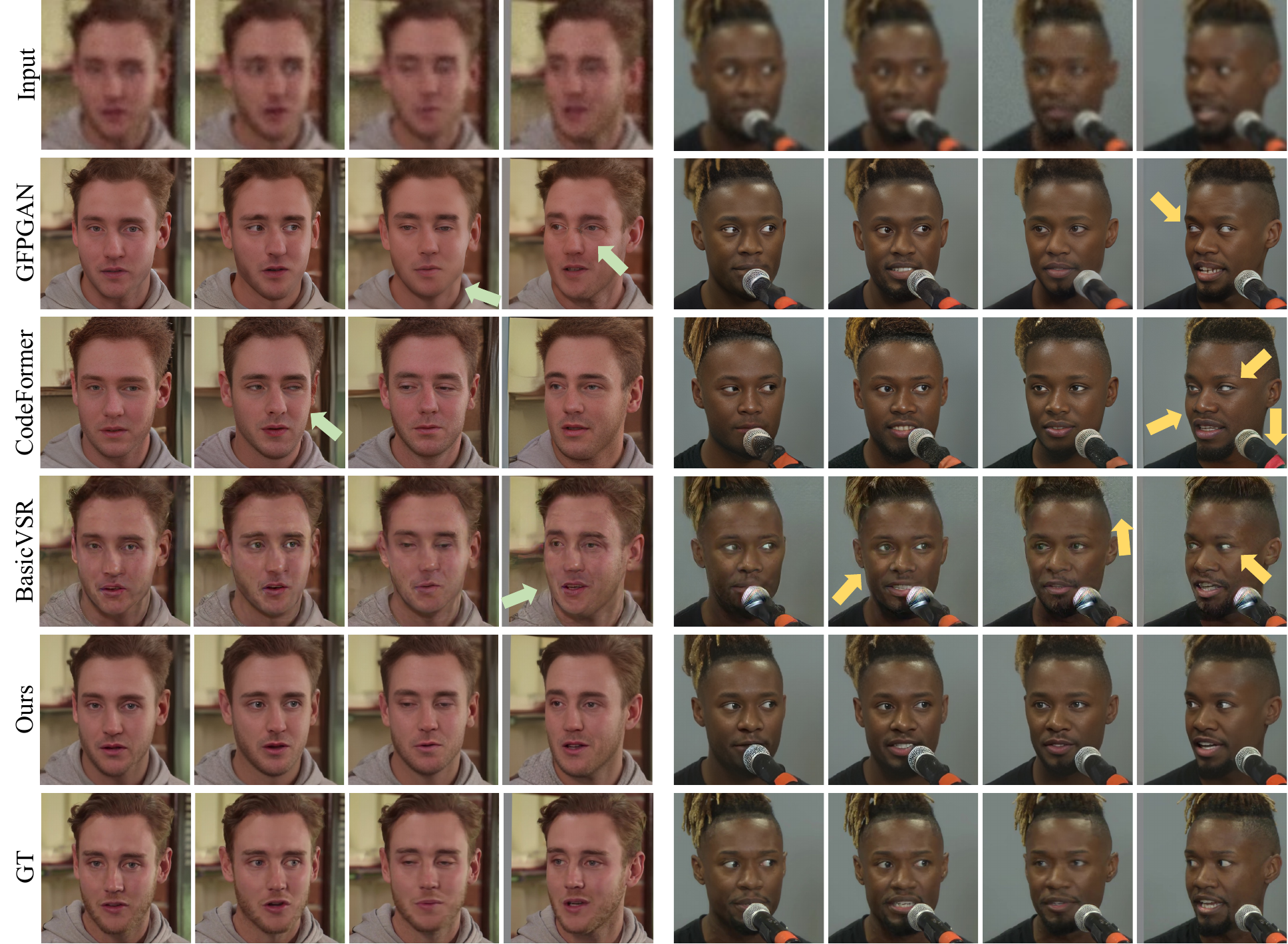}
    % \vspace{-0.2cm}
    \caption{
        \textbf{Qualitative comparison on the VFHQ.}
        Our KEEP produces high-fidelity face videos with faithful and consistent details. See arrows for details.}
    \label{fig:sota_comp}
    % \vspace{-0.2cm}
\end{figure*}

\emp{Quantitative Evaluation.}
The quantitative results are listed in Table~\ref{tab:sota_comparison}.
We observe that our method achieves better results than existing methods across all the metrics.
The results indicate that KEEP can faithfully recover facial details while preserving the identity.
Our method also maintains temporal coherence across frames, as quantified by $\sigma_{IDS}$ and $\sigma_{AKD}$, which represent the fluctuation of restored face identities and facial shapes.
Though exhibiting structural distortions and artifacts (See Fig.~\ref{fig:sota_comp}), general VSR models (EDVR, BasicVSR, BasicVSR++) typically achieve higher performance on fidelity metrics (PNSR, SSIM, LPIPS) than single-image FSR models. This suggests that image-based models produce high-quality but relatively low-fidelity results. Inconsistency could be introduced when latent code estimation is noisy and inaccurate.

\emp{Qualitative Evaluation.}
In Fig.~\ref{fig:sota_comp}, we can observe that the compared methods fail to reconstruct consistent appearances with perceptually pleasant details.
For example, GFPGAN tends to hallucinate facial details (\eg, ears in the second frame and incomplete glass in the last frame of the left example).
CodeFormer produces unnatural facial shapes (\eg, eyes), and BasicVSR leaves severe artifacts on the face images. 
In the last frame of the right example, both GFPGAN and CodeFormer generate unpleasant eyes (see yellow arrows).
In contrast, our \METHODNAME~exploits temporal information and restores finer and coherent facial details.
We refer readers to supplementary files for more video results.

\subsection{More Analysis}
\emp{Effectiveness of KFN.}
We first investigate the effectiveness of the Kalman Filter Network (KFN). As shown in Table~\ref{tab:ablation}, removing KFN results in worse performance on LPIPS, IDS, and AKD.
The results suggest the design of KFN is the key to promoting temporal consistency and identity preservation.

\emp{Effectiveness of CFA.}
Table~\ref{tab:ablation} also shows that Cross-Frame Attention (CFA) can further improve the performance.
Though not clearly reflected in the number, we further qualitatively show the effectiveness of KFN and CFA in the supplementary video. From the demo video, we can observe that 1) adopting KFN can ensure consistency in global style and maintain the global appearance of the recovered faces. 2) adding CFA can further coherently render local texture details (\eg, hair), and suppress flicker.
% \begin{table}
%     \centering
%     \caption{Ablation study of variant networks.}
%     \label{tab:ablation_cfa}
%     \vspace{-0.1cm}
%     \begin{tabular}{c|ccc}
%     \toprule
%         Metrics & LPIPS$\downarrow$ & IDS$\uparrow$ & AKD$\downarrow$ \\
%     \midrule
%         w/o KFN & 0.1721    & 0.7773    & 9.1952 \\
%         w/ KFN & 0.1619 & 0.7960    & 8.8182 \\
%     \bottomrule
%     \end{tabular}
%     \vspace{-0.3cm}
% \end{table}

\begin{table}[t]
    \centering
    \begin{minipage}[t]{.48\linewidth}
      \caption{Ablation study of variant networks.}
      \label{tab:ablation}
      % \vspace{-0.3cm}
      \centering
      \resizebox{.8\linewidth}{!}{
        \begin{tabular}{c|ccc}
            \toprule
                Models & LPIPS$\downarrow$ & IDS$\uparrow$ & AKD$\downarrow$ \\
            \midrule
                w/o CFA & 0.1621    & 0.7970    & 8.9029 \\
                w/o KFN & 0.1721    & 0.7773    & 9.1952 \\
                Full model & 0.1619 & 0.7960    & 8.8182 \\
            \bottomrule
        \end{tabular}
       }
    \end{minipage}
    \hfill
    \begin{minipage}[t]{.48\linewidth}
      \caption{Ablation study on optical flow estimator.}
      \label{tab:ablation_flow}
      % \vspace{-0.3cm}
      \centering
      \resizebox{.9\linewidth}{!}{
        \begin{tabular}{c|ccc}
            \toprule
                Estimator & LPIPS$\downarrow$ & IDS$\uparrow$ & AKD$\downarrow$ \\
            \midrule
                PWC-Net~\cite{sun2018pwc}   & 0.1623    & 0.7957    & 8.7839 \\
                GMFlow~\cite{xu2022gmflow} (Ours)               & 0.1619    & 0.7960    & 8.8182 \\
            \bottomrule
        \end{tabular}
    }
    \end{minipage}
    % \vspace{-0.3cm}
\end{table}

\emp{Effectiveness of Various Flow Estimator.}
We compare models with different flow estimators in Table~\ref{tab:ablation_flow}. The results suggest that the accuracy of estimated flows does not significantly affect the final performance.
We conjecture this can be attributed to two factors:
1) Minor misalignment in pixel space can be reasonably diminished as the latent code is highly downsampled by a factor of $32\times$, at which level the latent representations are less sensitive to small spatial discrepancies present in the pixel space.
2) Other modules can compensate for the inaccuracy caused by flow estimators in a joint training fashion.

\begin{figure*}[t]
    \centering
    \includegraphics[width=.9\linewidth]{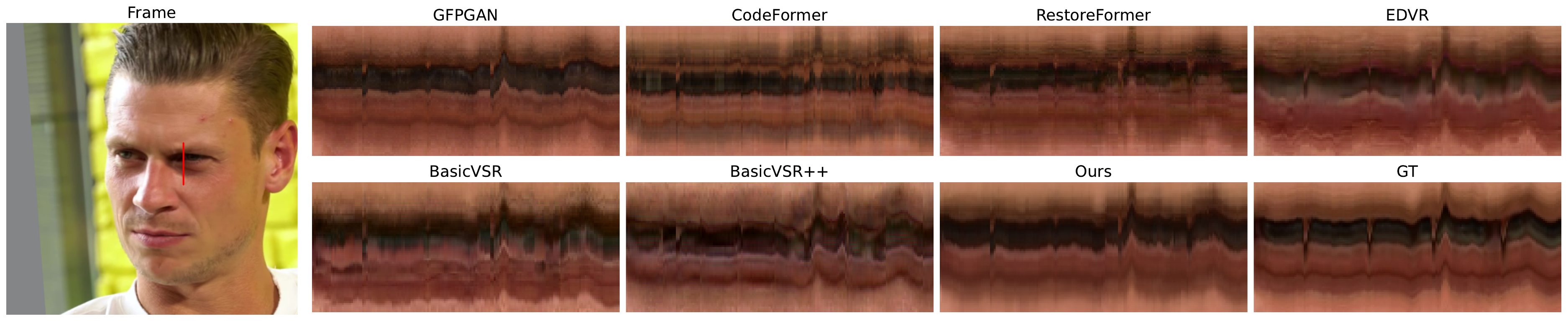}
    \includegraphics[width=.9\linewidth]{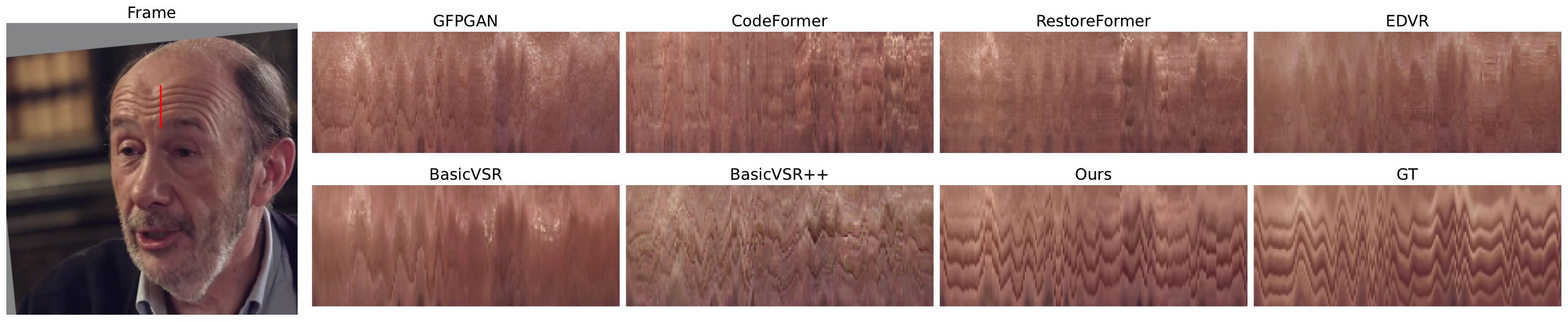}
    % \vspace{-0.2cm}
    \caption{
        \textbf{Comparison of temporal flicker.}
        We select each frame's column (red lines) and show the changes across time.
        Image-based models (GFPGAN, CodeFormer, and RestoreFormer) have obvious discontinuity around the eyes and wrinkles, and general VSR methods leave artifacts behind.
        In contrast, by maintaining stable facial priors and aggregating temporal information, our method remarkably suppresses temporal jitters and promotes coherent local details.
    }
    \label{fig:flicker}
    % \vspace{-.2cm}
\end{figure*}
\begin{figure}[t]
  % \vspace{-0.4cm}
    \centering
    \includegraphics[width=.9\linewidth]{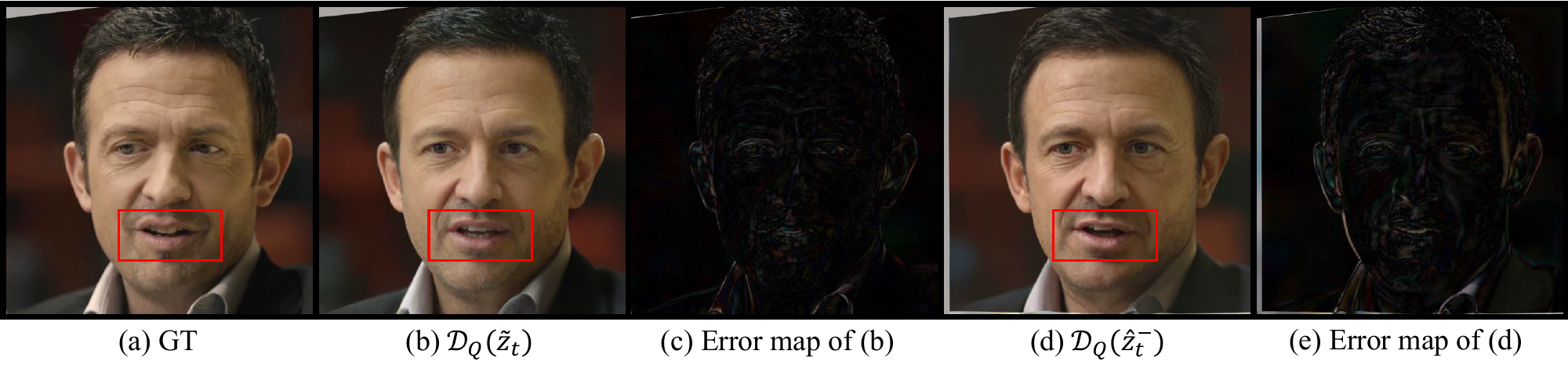}
      % \vspace{-0.7cm}
    \caption{
        Illustration of predicted state and observed state through Decoder $\mathcal{D}_Q$.
        The teeth in red boxes illustrate a slight difference between $\Tilde{\vb*{z}}_t$ and $\hat{\vb*{z}}_t^-$ when decoded in pixel space, suggesting the potential to supplement and fuse information to obtain a stable latent code.
    }
    \label{fig:latent_space}
  % \vspace{-0.5cm}
\end{figure}

\emp{Analysis of Flickering.}
We extract a short vertical segment of pixels from each frame and stack them horizontally to visualize the jittering issues within the video.
In particular, existing methods demonstrate clear jitter and flicker across time, while our method shows better temporal consistency.
Fig.~\ref{fig:flicker} demonstrates that other image-based models bring obvious jitters around eyes, and general VSR methods leave behind artifacts, while our method could remarkably suppress temporal jitters and promote coherent local details.

\emp{Analysis of Latent Space.}
Since the true state $\vb*{z}_t$ is unavailable, we indirectly analyze the predicted state $\Tilde{\vb*{z}}_t$ and observed state $\hat{\vb*{z}}_t^-$ by decoding them to the pixel space through Decoder $\mathcal{D}_Q$.
As shown in Fig. \ref{fig:latent_space}, the areas around teeth exhibit a slight difference, while most remaining parts in error maps show similar decoded results.
This indicates that the predicted and observed states could supplement each other to obtain a more accurate estimation of $\vb*{z}_t$, which is where the power of Kalman filter lies.

\begin{figure}[t]
    \centering
    \includegraphics[width=.86\linewidth]{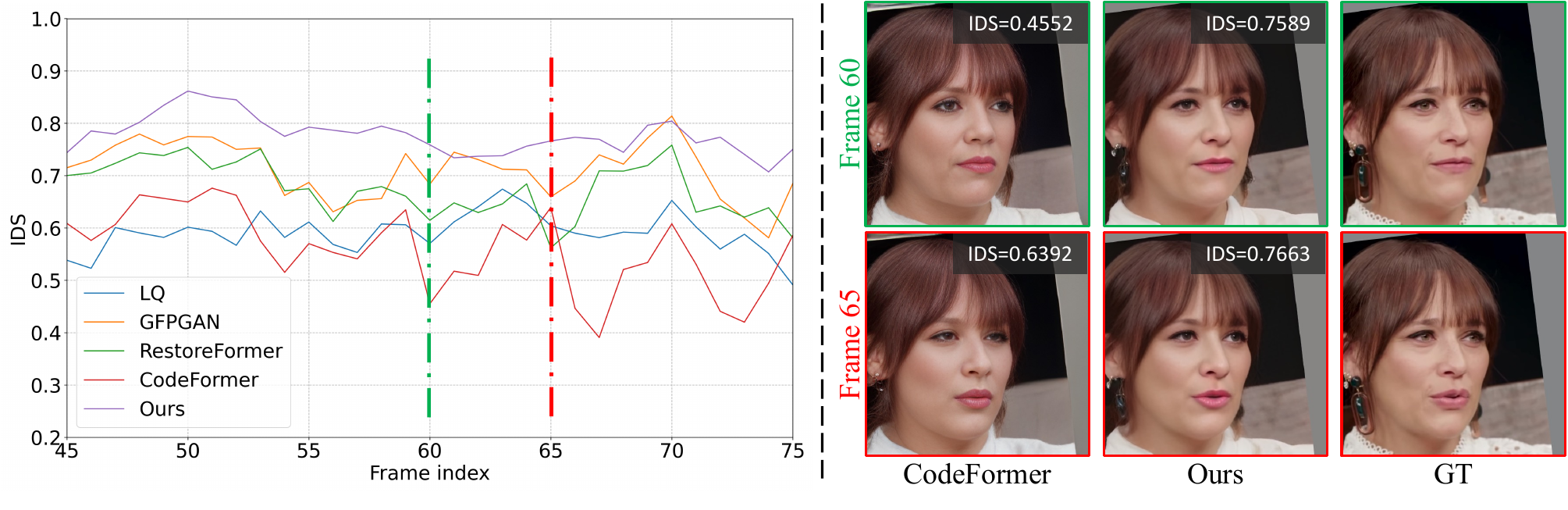}
    % \vspace{-0.2cm}
    \caption{\textbf{Identity similarity across frames.}
    Our method preserves the identity of input images and exhibits less fluctuation over time.
    CodeFormer results of two frames (green and red dashed line) exemplify an abrupt change of identity in the right figure, while our method maintains a stable identity both quantitatively and qualitatively.
    % \cavan{Sorry to say this but I don't think you have paid much effort to draw a nice figure here. It just takes too much space, but just conveys a very simple message. Did you remember what we discussed last time?}
    }
    \label{fig:sim}
    % \vspace{-0.2cm}
\end{figure}
\begin{figure*}[t]
    \centering
    \includegraphics[width=.88\linewidth]{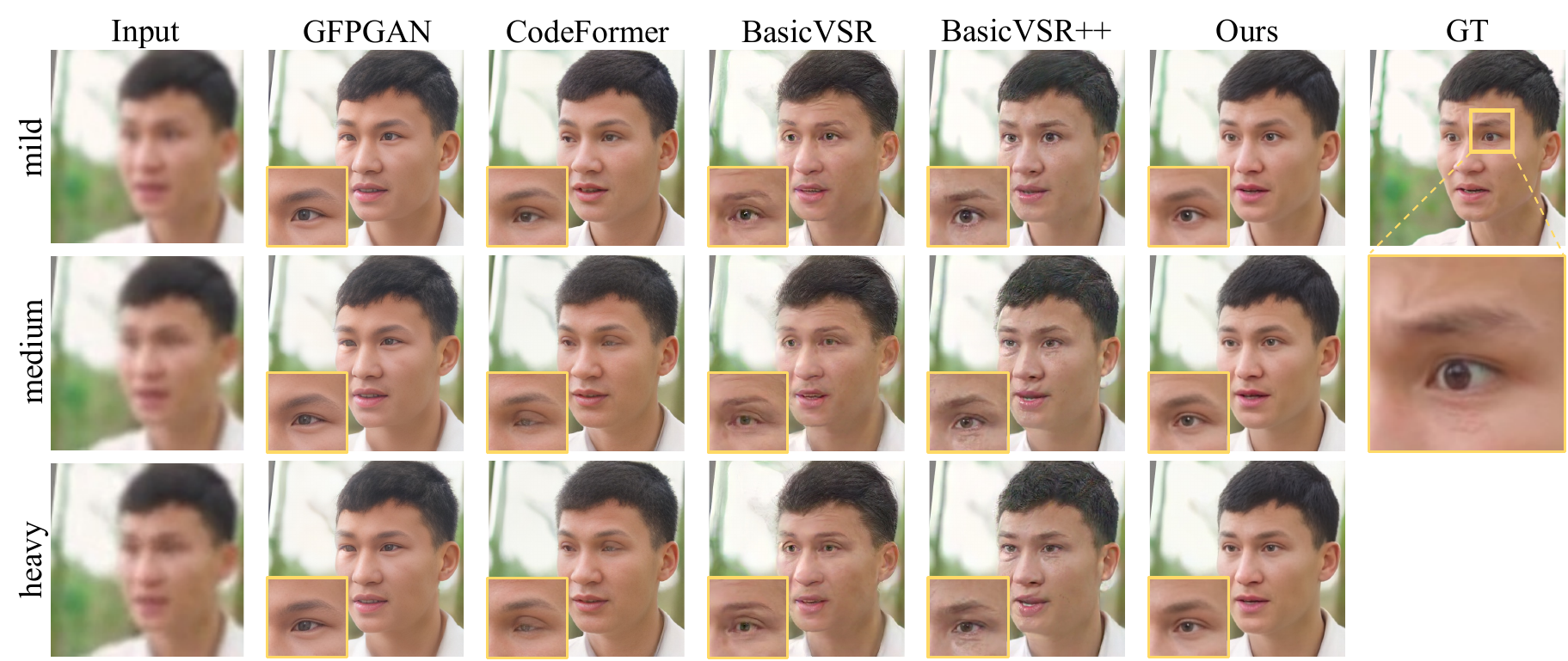}
    % \vspace{-0.2cm}
    \caption{
        \textbf{Qualitative comparison on different levels of degradation.}
        Our KEEP maintains high-fidelity in various degradations.
        % \cavan{Can't see a huge difference between the mild, medium and heavy input.}
        }
    \label{fig:various_degrdation}
    % \vspace{-0.5cm}
\end{figure*}
\begin{figure}[ht]
     \centering
     \begin{subfigure}[b]{0.48\textwidth}
         \centering
        \includegraphics[width=.98\linewidth]{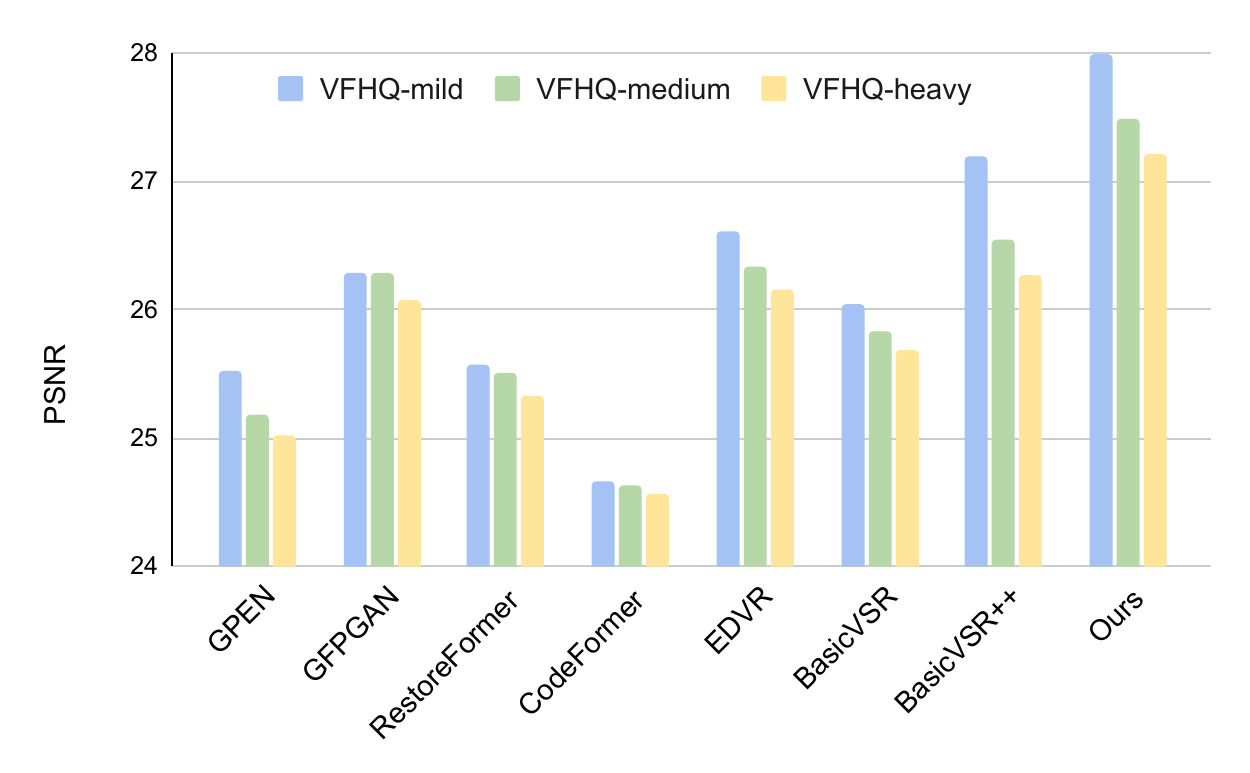}
        % \vspace{-0.2cm}
        \includegraphics[width=.98\linewidth]{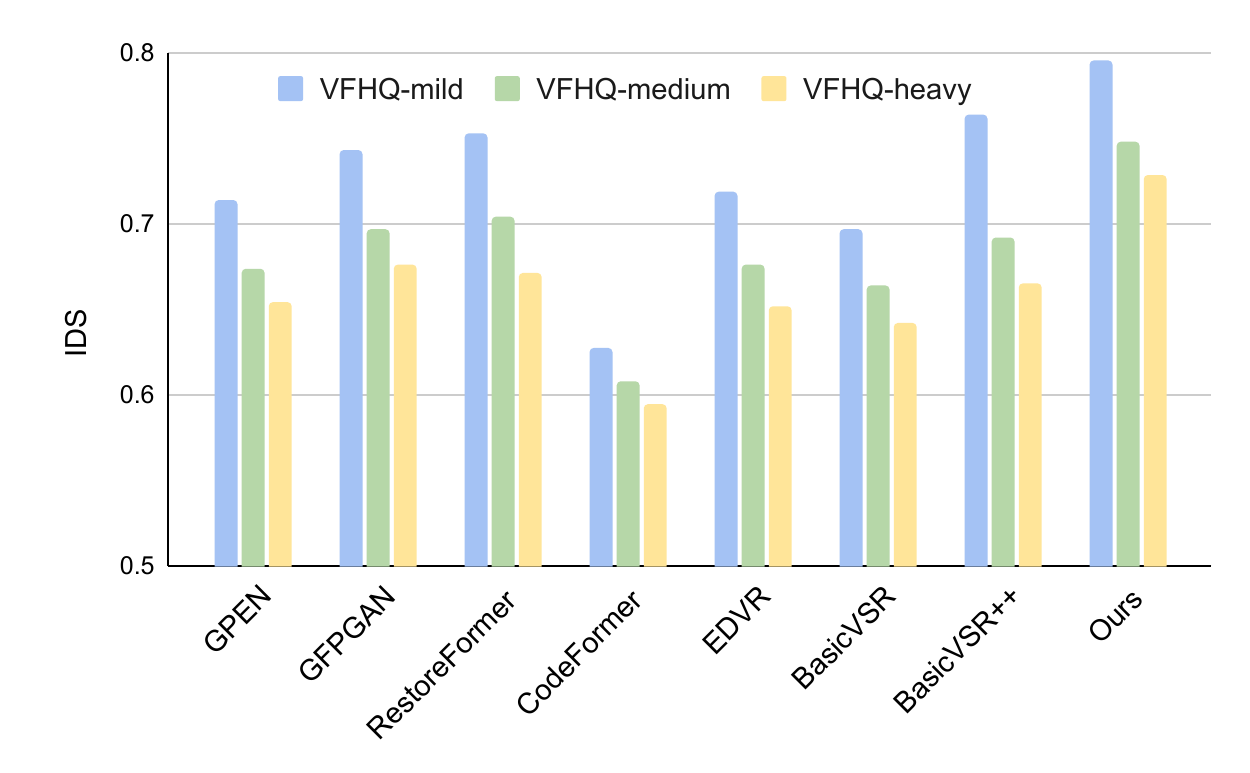}
        % \vspace{-0.2cm}
        \caption{PNSR and IDS on different partitions.}
        \label{fig:bar_var}
     \end{subfigure}
     \hfill
     \begin{subfigure}[b]{0.51\textwidth}
         \centering
         \includegraphics[width=.98\linewidth]{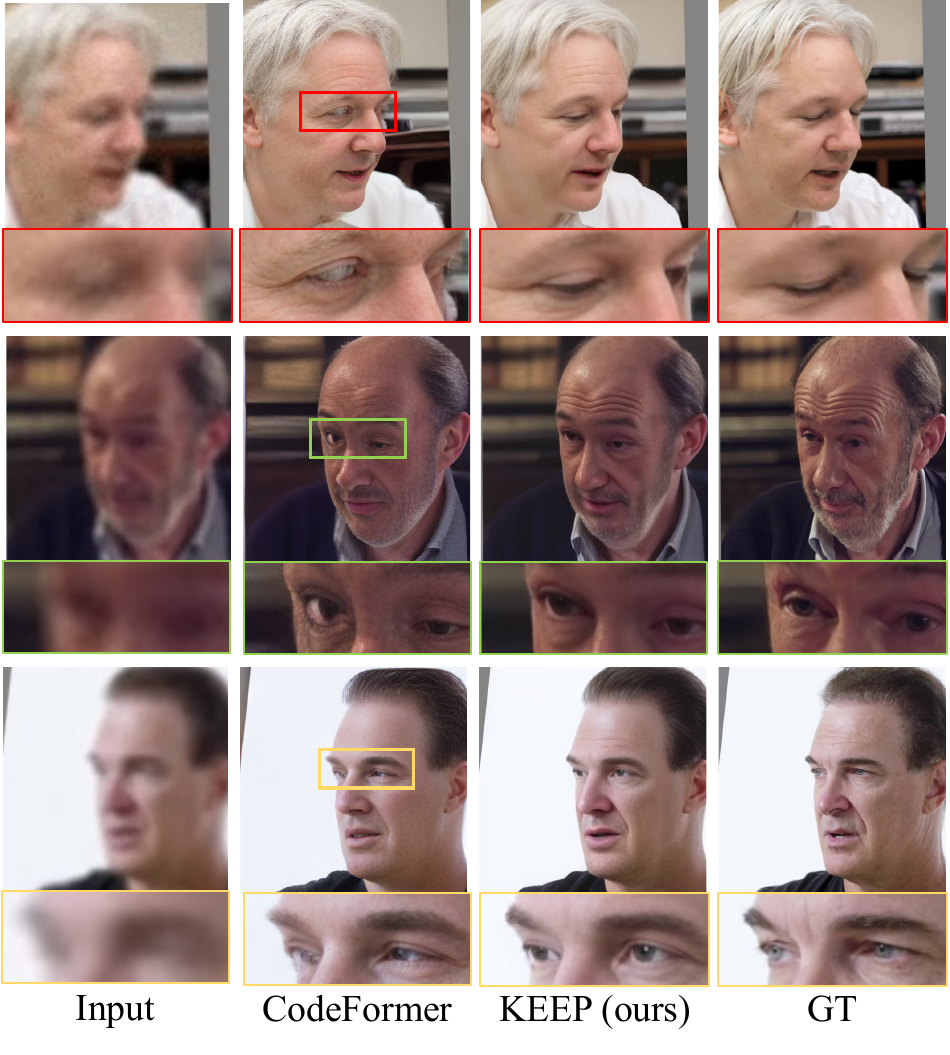}
         \caption{Comparisons on non-frontal faces.}
         \label{fig:non_frontal}
     \end{subfigure}
    \caption{
    (a) Our methods achieve consistently better performance on various levels of degradation.
    (b) While CodeFormer fails to restore eyes to these challenging cases, our method can still produce plausible facial elements. }
    \label{fig:combo1}
    % \vspace{-0.2cm}
\end{figure}

\emp{Identity Preservation.}
We show the identity similarity across frames of one representative video clip in Fig.~\ref{fig:sim} 
As illustrated in the left figure, our method achieves better identity preservation and less identity jitter within the video, compared to the single-image methods.
We also exemplify that the identity of CodeFormer results can change abruptly within several frames. The identity score of CodeFormer increases from $0.4552$ to $0.6392$, and turns down in later frames (See red curve), demonstrating great fluctuation across time.
In contrast, our method maintains a stable identity both quantitatively and qualitatively.

\emp{Various Degradations.}
Fig.~\ref{fig:bar_var} shows that KEEP is consistently better than the compared methods across different difficulty levels.
Fig.~\ref{fig:various_degrdation} demonstrates results on different levels of degradation.
We draw the following observations.
1) Even GFPGAN and CodeFormer can restore plausible results on frames with mild degradation, the performance significantly deteriorates (\eg, eyes) upon heavier degradation.
2) Our method achieves better results by considering complementary information between adjacent frames and maintaining stable face priors.
In addition, our method is appealing in handling heavy degradation.

\emp{Non-Frontal-View Faces.}
In Fig.~\ref{fig:non_frontal}, our model shows enhanced performance on non-frontal faces by providing more stable face priors estimations. While the single-image model CodeFormer cannot recover the eyes, our KEEP is still able to show robustness to these challenging cases.
% \input{figure/non_frontal}

% \input{figure/bar_var}

% \emp{Analysis of Kalman Gain.}
% Histograms of gains in three representative video sequences are shown in Fig. \ref{fig:histogram}, revealing a distribution primarily within range [0, 0.4].
% This implies the predicted states complement the new measured states, demonstrating that they are not converging to trivial solutions. \cavan{Perhaps move the detailed histogram to the supplementary and report the overall statistics on more sequences here.}
% \input{figure/histogram}

\section{Conclusion}
We present a novel framework, \METHODNAME, aiming at resolving the challenges associated with facial detail and temporal consistency in video face restoration.
The proposed method demonstrates a unique capability to maintain a stable face prior over time, which is achieved by Kalman filtering principles, where our approach recurrently incorporates information from previously restored frames to guide and regulate the restoration process of the current frame.
Extensive experiments demonstrate the efficacy of \METHODNAME\ in consistently capturing facial details across video frames and keeping the temporal stability of face videos.

\section*{Acknowledgment}
This study is supported under the RIE2020 Industry Alignment Fund Industry Collaboration Projects (IAF-ICP) Funding Initiative, as well as cash and in-kind contribution from the industry partner(s).

% \clearpage  % TODO REVIEW/FINAL: This \clearpage needs to be removed from both review and camera-ready versions.

% ---- Bibliography ----
%
% BibTeX users should specify bibliography style 'splncs04'.
% References will then be sorted and formatted in the correct style.
%
\bibliographystyle{splncs04}
\bibliography{egbib}
\end{document}

% ---------------------------------------------------------------
% TODO REVIEW: Replace with your title
\title{Kalman-Inspired Feature Propagation for \\ Video Face Super-Resolution\\
- Supplementary Materials -}

% TODO REVIEW: If the paper title is too long for the running head, you can set
% an abbreviated paper title here. If not, comment out.
\titlerunning{KEEP}

% TODO FINAL: Replace with your author list. 
% Include the authors' OCRID for the camera-ready version, if at all possible.
\author{Ruicheng Feng\orcidlink{0000-0003-4544-3078} \and
Chongyi Li\orcidlink{0000-0003-2609-2460} \and
Chen Change Loy\orcidlink{0000-0001-5345-1591}}

% TODO FINAL: Replace with an abbreviated list of authors.
\authorrunning{R.~Feng et al.}
% First names are abbreviated in the running head.
% If there are more than two authors, 'et al.' is used.

% TODO FINAL: Replace with your institution list.
\institute{S-Lab, Nanyang Technological University, Singapore\\
\email{\{ruicheng002, ccloy\}@ntu.edu.sg}\\
\email{lichongyi25@gmail.com}
}

\maketitle
\appendix

\setcounter{table}{0}
\renewcommand{\thetable}{A\arabic{table}}
\setcounter{figure}{0}
\renewcommand{\thefigure}{A\arabic{figure}}

\section{Method Details}
\subsection{Detailed Architecture}
\begin{figure*}[h]
    \centering
    \includegraphics[width=.98\linewidth]{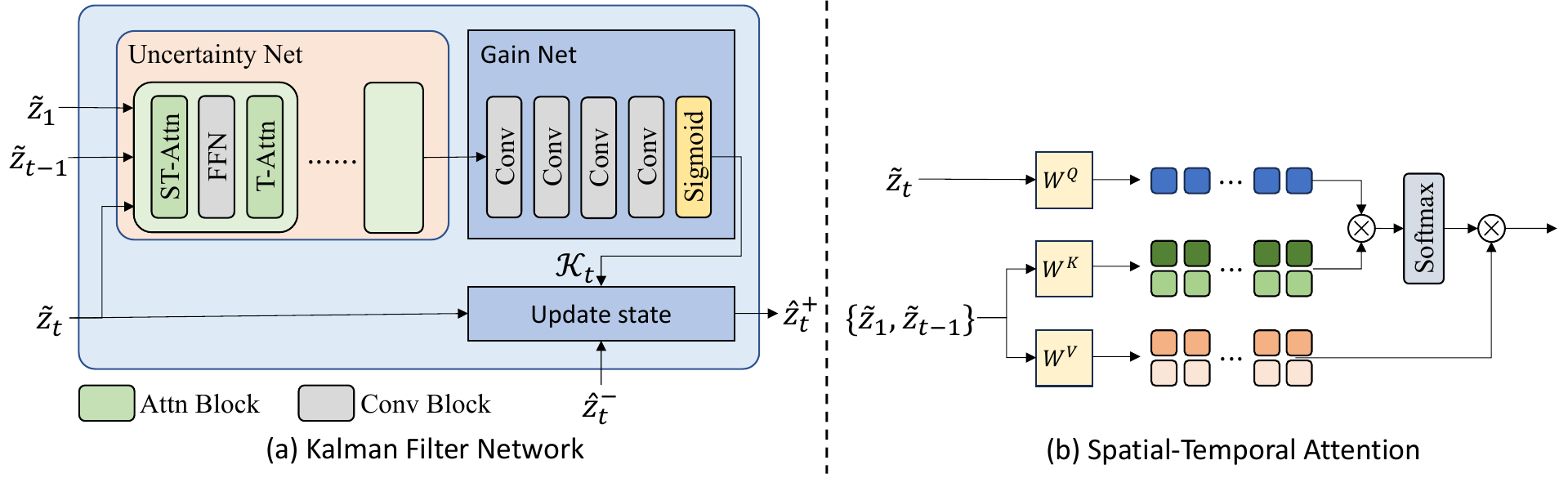}
    % \vspace{-0.2cm}
    \caption{
        \textbf{Illustration of Kalman Filter Network.}
        (a) We unfold and show one timestep of the Kalman filter network. The network mainly consists of two parametrization modules, \ie, uncertainty network and the gain network. Here ``ST-Attn'' and ``T-Attn'' represent spatial-temporal attention and temporal attention, respectively.
        (b) The Spatial-Temporal Attention (ST-Attn) takes estimated observed latent code for current frame $\Tilde{\vb*{z}}_t$ as a query and attends to the combination of the first frame $\Tilde{\vb*{z}}_1$ and previous frame $\Tilde{\vb*{z}}_{t-1}$.  
    }
    \label{fig:kalman_filter_arch}
    % \vspace{-0.5cm}
\end{figure*}

\paragraph{\textbf{Kalman Filter Network.}}
Our Kalman Filter Network, as illustrated in Figure~\ref{fig:kalman_filter_arch}, adopts two distinct parameterized modules in implicitly estimating uncertainty and Kalman gain.
Note that the dynamic model from $\hat{\vb*{z}}_{t-1}^+$ to prior estimation $\hat{\vb*{z}}_{t}^-$ is omitted for simplicity in the illustration.
The uncertainty network implicitly estimates the uncertainty of shape $h\times w \times c$, and the Kalman gain network calculates the corresponding Kalman gain $\mathcal{K}_t$ of shape $h\times w$ for each code token.
The Spatial-Temporal Attention (ST-Attn) takes the current observed latent code $\Tilde{\vb*{z}}_t$ as a query and attends to the combination of the first frame $\Tilde{\vb*{z}}_1$ and previous frame $\Tilde{\vb*{z}}_{t-1}$. 
Inspired by \cite{wu2023tune}, the spatial-temporal attention also takes the latent code of the first frame $\Tilde{\vb*{z}}_1$ as input, which serves as an anchor prior to all temporal attention.

\begin{figure}[!t]
    \centering
    \includegraphics[width=.98\linewidth]{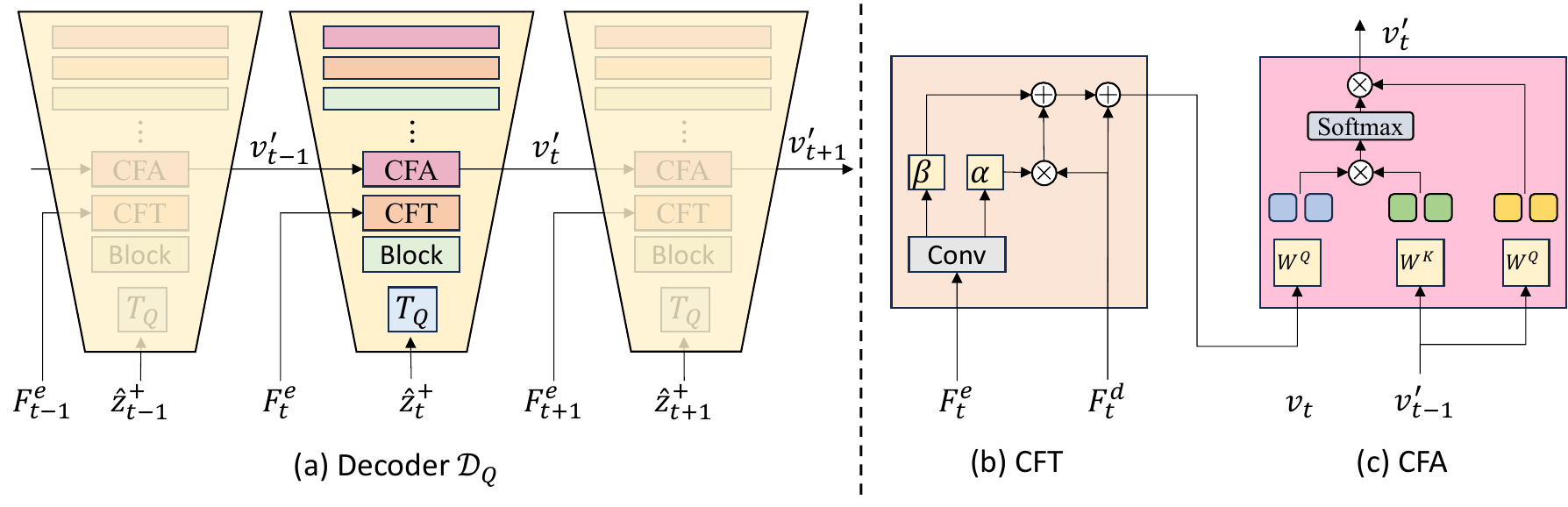}
    % \vspace{-0.2cm}
    \caption{
        \textbf{Illustration of the integrated decoder.}
        Controllable Feature Transformation (CFT) and Cross-Frame Attention (CFA).
        $T_Q$ is a codebook lookup Transformer and quantization layer borrowed from CodeFormer~\cite{zhou2022codeformer}.
        Blocks are the basic conv blocks in the decoder.
        CFT is tailored for modulating the features of decoder $F_d$ by the encoder's features $F_e$.
        CFA is adopted in the decoder to further promote local temporal consistency to regularize the information propagation.
    }
    \label{fig:decoder_arch}
    % \vspace{-0.5cm}
\end{figure}

\paragraph{\textbf{Integrated Decoder.}}
Figure \ref{fig:decoder_arch} depicts how CFT and CFA layers are integrated into the decoder.
Following \cite{zhou2022codeformer}, we leverage the encoder features to modulate the corresponding decoder features.
Denoted $F_e$ and $F_d$ as the encoder and decoder features, respectively, the network learns an affine transformation defined by $\alpha$ and $\beta$.

\begin{equation}
    v_t = F_d + (\alpha \cdot F_d + \beta),
\end{equation}
where $\alpha, \beta = \mathcal{C}(F_e)$, and $\mathcal{C}$ is multiple convolution blocks.
The CFT modules are adopted at multiple scales ${16, 32, 64}$, since shallow features of encoder would also bring forward corrupted information to the decoder and yield blurry results.
This design facilitates fidelity reservation of each frame and hence improves temporal coherence.

To further enforce temporal information propagation and reduce local jitters, we adopt cross-frame attention modules, which search and match similar features from the previous frame and attend to them correspondingly.
Specifically, given the latent features from the previous frame $v_{t-1}$ and current frame $v_t$. They are projected onto the embedding space and output the features $v_i'$ by
$v_i'=\text{Attn}(Q, K, V)=\text{softmax}(\frac{QK^T}{\sqrt{d}})\cdot V$, where
\begin{equation}
    Q=W_Q\cdot v_t, K=W_K\cdot v_{t-1}, V=W_V\cdot v_{t-1}.
\end{equation}
This module facilitates temporal information propagation in the decoder.
We adopt CFA modules on features of small scale $16$ and $32$ to avoid introducing blur to the decoded results.

\SetKwComment{Comment}{/* }{ */}
\begin{algorithm}[!t]
    \caption{Detailed algorithm of KEEP.}
    \label{alg}
    $\mathcal{E}_{L}, \mathcal{E}_{H}, \mathcal{D}_Q \gets \text{LQ Encoder / HQ Encoder/ Decoder}$\;
    $\varphi \gets \text{Kalman Gain network}$\;
    $\Phi_{t-1\rightarrow t} \gets \text{Optical flow from previous frame}$\;
    $\omega \gets \text{Spatial warping operation}$\;
    $T \gets \text{length of clips}$\;
    \textbf{Initialize} $\Tilde{\vb*{z}}_1$, $\hat{\vb*{z}}_1^+$, $\hat{\vb*{y}}_1$\;
    \For{$t = 2, 3,\cdots,T$}{
        \textbf{State Prediction:}\\
            $\hat{\vb*{z}}_t^- \gets \mathcal{E}_H(\omega(\hat{\vb*{y}}_{t-1}, \Phi_{t-1\rightarrow t}))$\;
        \textbf{State Update:}\\
            $\Tilde{\vb*{z}}_t \gets \mathcal{E}_{L}(\vb*{x}_t)$\;
            $\mathcal{K}_t \gets \varphi (\Tilde{\vb*{z}}_1, \Tilde{\vb*{z}}_{t-1}, \Tilde{\vb*{z}}_t)$\;
            $\hat{\vb*{z}}_t^+ \gets (1-\mathcal{K}_t) \hat{\vb*{z}}_t^- + \mathcal{K}_t \Tilde{\vb*{z}}_t$\;
            $\hat{\vb*{y}}_t \gets \mathcal{D}_Q(\hat{\vb*{z}}_{t}^+))$
    }
\end{algorithm}

\subsection{Algorithm Pseudocode}
As shown in the below Algorithm \label{alg}, we present the pseudocode of our method.
In this algorithm, we only show the process of inference.

\subsection{Training Scheme.}
\paragraph{\textbf{Codebook Pre-Training (Stage I).}}
Following CodeFormer \cite{zhou2022codeformer}, we first pre-train a codebook within a quantized autoencoder. Unlike TAST~\cite{ge2022long}, the learned codebook is still image-based and does not involve temporal information.
Precisely, given a HQ frame $\vb*{y}_t\in\mathbb{R}^{H\times W\times 3}$ in pixel space, an encoder in HQ domain $\mathcal{E}_H$ encodes it into a latent code $\mathcal{E}_{H}(\vb*{y}_t)$.
Each token of the continuous code will be mapped to quantized discrete code $\hat{\vb*{z}}_t^q$ from the learnable codebook $\mathcal{C}=\{c_k\in \mathbb{R}^d\}^N_{k=0}$ via nearest-neighbor matching.
The decoder $\mathcal{D}$ then reconstructs the high-quality image frame from latent code.
Similar to \cite{zhou2022codeformer,esser2021taming},
to train the quantized autoencoder, we adopt three image-level reconstruction losses: pixel loss$\mathcal{L}_1$, perceptual loss \cite{johnson2016perceptual,zhang2018unreasonable} $\mathcal{L}_{per}$, and adversarial loss \cite{wang2018esrgan} $\mathcal{L}_{adv}$.
Moreover, since image-level losses are underconstrained when updating the discrete codebook, code-level losses are also used to reduce the distance between the quantized code and input feature embeddings. The overall objectives in this stage are defined by
\begin{equation}
    \mathcal{L}_I = \mathcal{L}_1+\mathcal{L}_{per}+\mathcal{L}_{adv}+||sg(\mathcal{E}_{H}(\vb*{y}_t))-\hat{\vb*{z}}_t^q||^2_2+||\mathcal{E}_{H}(\vb*{y}_t)-sg(\hat{\vb*{z}}_t^q)||^2_2,
\end{equation}
where $sg(\cdot)$ denotes stop-gradient operator.

\paragraph{\textbf{Kalman Filter Network(Stage II).}}
In this stage, we train a LQ encoder $\mathcal{E}_L$, the quantization Transformer $T_q$, and the Kalman filter network, while the codebook $\mathcal{C}$ and decoder $\mathcal{D}$ are frozen to preserve high-quality restoration from the VQGAN.
Similar to \cite{zhou2022codeformer}, we adopt cross-entropy loss $\mathcal{L}_{CE}$ to supervise token prediction, and feature loss $\mathcal{L}_2$ to minimize the distance between features before and after quantization. The overall objectives are defined by
\begin{equation}
    \mathcal{L}_{II} = \mathcal{L}_{CE}+\mathcal{L}_2(\mathcal{E}_{H}(\vb*{y}_t), sg(\hat{\vb*{z}}_t^q))).
\end{equation}

\paragraph{\textbf{Cross-Frame Attention (Stage III).}}
To train both Cross-Frame Attention (CFA) modules and Controllable Feature Transformation (CFT),
we fix other modules and use image-level reconstruction loss $\mathcal{L}_1$, $\mathcal{L}_{per}$ and GAN loss $\mathcal{L}_{adv}$, given by
\begin{equation}
    \mathcal{L}_D = E[\text{log} D(Y)] + E[1 - \text{log} D(\hat{Y})].
\end{equation}
The discriminator $D$ is constructed with multiple 3D convolution layers \cite{chang2019free}, denoted as temporal PatchGAN, which could further enhance the coherence of the generated face videos.
The adversarial loss for the decoder modules is formulated as
\begin{equation}
    \mathcal{L}_{adv} = - E[\text{log} D(\hat{Y})].
\end{equation}
Additionally, we adopt temporal loss\cite{lai2018learning} between consecutive output frames, formulated as
\begin{equation}
    \mathcal{L}_{temp}=\sum_{t=2}^T M_{t-1\rightarrow t}\cdot||\hat{\vb*{y}}_t - \hat{\vb*{y}}_{t-1\rightarrow t}||_1,
\end{equation}
where $M_{t-1\rightarrow t}$ denotes the valid mask computed by forward-backward consistency assumption~\cite{sundaram2010dense}, and $\hat{\vb*{y}}_{t-1\rightarrow t}$ is the frame warped from previous frame $\hat{\vb*{y}}_{t-1}$ with optical flow estimated by GT frames $\vb*{y}_{t-1}$ and $\vb*{y}_{t}$.

Hence, the overall training objectives are given by
\begin{equation}
    \mathcal{L}_{III} = \lambda_1\mathcal{L}_{1} + \lambda_{per}\mathcal{L}_{per} + \lambda_{adv}\mathcal{L}_{adv} + \lambda_{temp}\mathcal{L}_{temp}.
\end{equation}
Here $\lambda_1$, $\lambda_{per}$, $\lambda_{adv}$, and $\lambda_{temp}$ are the balancing weights and we empirically set $\lambda_1=0.01$, $\lambda_{per}=1$, $\lambda_{adv}=0.1$, and $\lambda_{temp}=0.1$.

\subsection{Details of Dataset}
Different from image-based degradations, video compression implicitly considers the dependencies across video frames, hence inducing temporal-variant degradations. This is implemented by randomly selecting codecs and constant rate factor (CRF) during training.
The overall degradation model is defined by 
\begin{equation}
    \vb*{x}=\{[(\vb*{y} \circledast \vb*{k}_\sigma)\downarrow + \vb*{n}_\delta]_{codec}\}\uparrow,
\end{equation}
where $\vb*{x}$ and $\vb*{y}$ are degraded and high-quality video clips, respectively. 
$\vb*{k}$ and $\vb*{n}_\delta$ are Gaussian blur kernel and Gaussian noise specified by $\sigma$ and $\delta$, respectively. $\circledast$ denotes the convolution operation, and $\downarrow$ and $\uparrow$ represent $4\times$ downsample and upsample in this paper. Video compression $codec$ is selected from ``libx264'' and ``h264'' and the video quality is controlled by CRF, ranging from [$25,45$]. During training, $\sigma$ is sampled from [$2, 10$], and noise level $\delta$ from [$0, 10$].

For a comprehensive evaluation, we synthesize three splits of the VFHQ-Test dataset containing different levels of degradation. As summarized in Table \ref{tab:dataset}, they follow the same degradation model but differ in the degree of noise, blur, and compression.
Note that we mainly focus on the video compression controlled by CRF, which is unique for video tasks.

Besides synthetic degradations, we also assess the generalizability of our methods on real-world face videos.
In particular, we collect $40$ videos in the wild from YouTube, covering various degradations and celebrities in different scenes, \eg, interviews, and talk shows.
Given the raw video from online sources, data processing pipeline proposed by \cite{xie2022vfhq} is adopted to extract low-quality real face videos.
For each video clip, we retain a sequence of $100$ to $300$ frames without scene transitions, which may break the dynamics between frames and hence deteriorate the temporal propagation.

\begin{table}
    \caption{We divide the test data into different levels of difficulty for a more comprehensive analysis.}
    \label{tab:dataset}
    % \vspace{-0.2cm}
    \centering
    \begin{tabular}{cccc}
    \toprule
       Degradation  & mild & medium & heavy \\
    \midrule
       Noise $\delta$  & [0, 5] & [5, 10] & [5, 10] \\
       Blur $\sigma$ & [2, 5] & [5, 10] & [5, 10] \\
       CRF  & [18, 25] & [25, 35] & [35, 45]\\
   \bottomrule
    \end{tabular}
    % \vspace{-0.4cm}
\end{table}

\section{More Analysis}
\subsection{Effectiveness of Alignment}
\begin{figure}[!t]
    \centering
    \includegraphics[width=.9\linewidth]{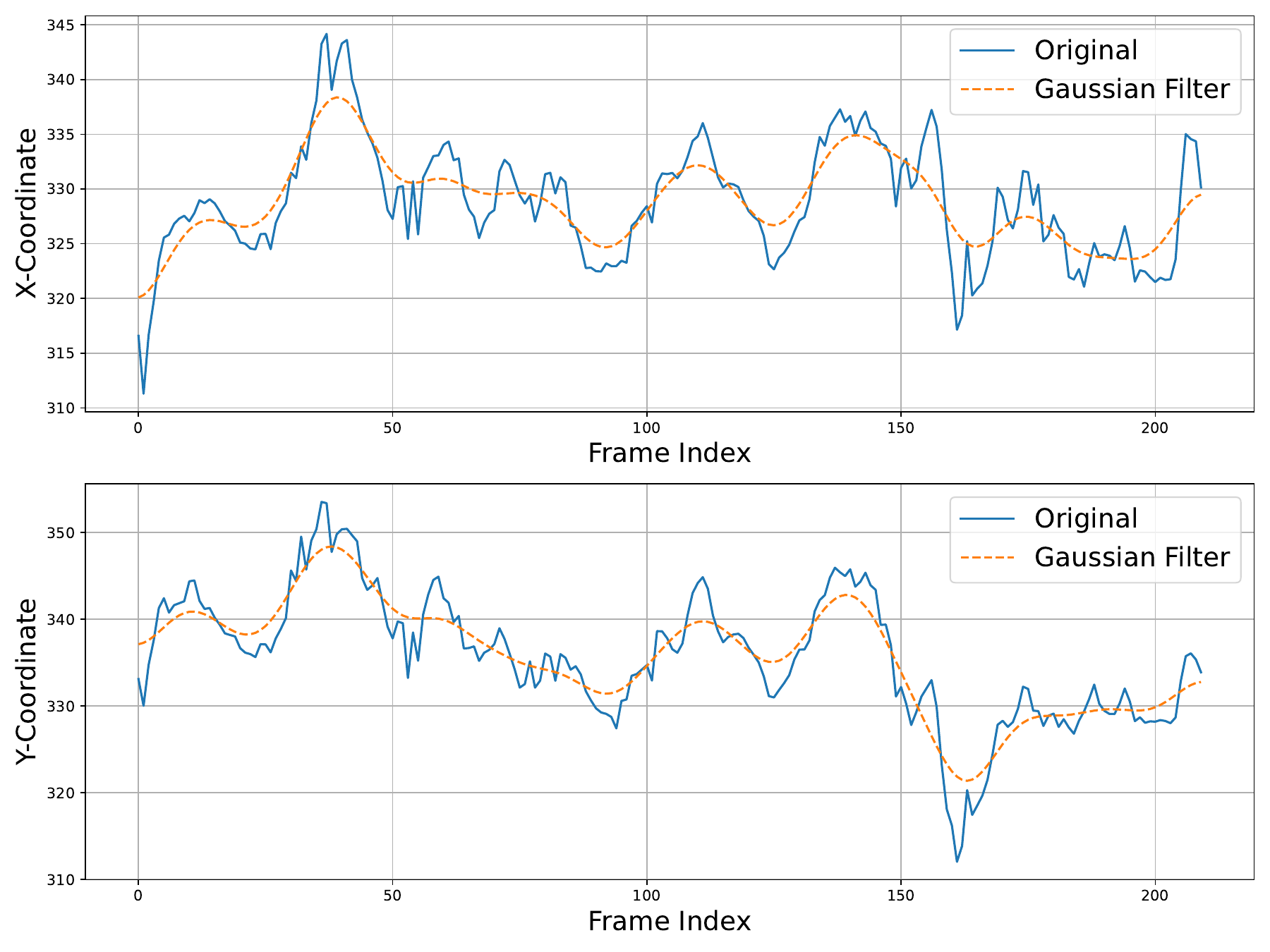}
    % \vspace{-0.2cm}
    \caption{
        A representative example of landmark location processed by Gaussian filter along time.
    }
    \label{fig:gaussian_filter}
    % \vspace{-0.5cm}
\end{figure}
Alignment is the common pre-processing procedure in face-related vision tasks. 
This ensures the faces are transformed and centralized in the same canonical coordinate system.
This is realized by detecting landmark keypoints and applying affine transformation to the original face images, which is sensitive to the locations of detected facial landmarks.
Mild inaccuracy of landmark detections is tolerable in a single image.
However, the noisy detections could consequently result in unintentional temporal inconsistencies between frames in a video.

To reduce the additional inconsistency, we adopt low-pass Gaussian filter on the locations of each landmarks along the temporal dimension, which eliminates abrupt change (jitters) oriented along time.
Denoted $\mathcal{M}_t^k$ as the $k$-th detected landmarks from frame $y_t$, where $t$ represents the timestep. The filtered landmarks are given by
\begin{equation}
    \hat{\mathcal{M}}_t^k = \sum_{n=t-r}^{t+r} G(n,\sigma)\cdot\mathcal{M}_n^k,
\end{equation}
where $G(n,\sigma)=\frac{1}{2\pi\sigma^2} e^{-\frac{(n-t)^2}{2\sigma^2}}$, and $r$ denotes the radius of the window size.
In our experiments, we empirically set $\sigma=5$ and $r=20$.
Figure~\ref{fig:gaussian_filter} provides a visual example of landmarks processed by Gaussian filter. The temporal jitters are largely alleviated by the filter.
We also demonstrate the effectiveness of alignment in the supplementary video.

\subsection{Quantitative Comparison on Various Degradation.}
\begin{table*}[t]
    \centering
    \caption{\textbf{Quantitative comparison on VFHQ dataset with different levels of degradation.} \red{Red} and \blue{Blue} indicate the best and the second best results.}
    \label{tab:full_sota}
    % \vspace{-0.1cm}
\begin{tabular}{l||ccc|cc|cc}
\toprule
Method        & \multicolumn{1}{l}{PSNR$\uparrow$} & \multicolumn{1}{l}{SSIM$\uparrow$} & \multicolumn{1}{l}{LPIPS$\downarrow$} & \multicolumn{1}{|l}{IDS$\uparrow$} & \multicolumn{1}{l|}{AKD$\downarrow$} & \multicolumn{1}{l}{$\sigma_{IDS}(\times 10^{-2})\downarrow$} & \multicolumn{1}{l}{$\sigma_{AKD}\downarrow$} \\
\midrule
              & \multicolumn{7}{c}{Mild}                                                                                                                                                                  \\
\midrule
GPEN \cite{yang2021gan}          & 25.5193                  & 0.7517                   & 0.2988                    & 0.7142                  & 11.4691                 & 4.7416                 & 3.5109                     \\
GFPGAN \cite{wang2021towards}        & 26.2933                  & 0.7795                   & 0.2482                    & 0.7437                  & 10.5467                 & \blue{4.5700}                 & 3.6482                     \\
RestoreFormer \cite{wang2022restoreformer} & 25.5720                  & 0.7344                   & 0.3195                    & 0.7530                  & \blue{10.5354}                 & 4.7159                 & \blue{3.4122}                     \\
CodeFormer \cite{zhou2022codeformer}    & 24.6597                  & 0.7454                   & 0.2742                    & 0.6272                  & 11.4983                 & 6.3726                 & 3.6927                     \\
EDVR \cite{wang2019edvr}          & 26.6051                  & 0.7858                   & 0.2484                    & 0.7195                  & 11.6220                 & 4.8048                 & 3.5829                     \\
BasicVSR \cite{chan2021basicvsr}      & 26.0458                  & 0.7765                   & 0.2496                    & 0.6973                  & 11.3679                 & 5.0343                 & 3.6054                     \\
BasicVSR++ \cite{chan2022basicvsr++}    & \blue{27.1996}                  & \blue{0.8057}                   & \blue{0.1958}                    & \blue{0.7641}                  & 11.3136                 & 5.2543                 & 4.6425                     \\
\textbf{KEEP (Ours)}          & \red{27.9994}                  & \red{0.8267}                   & \red{0.1619}                    & \red{0.7960}                  & \red{8.8182}                  & \red{3.6866}                 & \red{3.2538}                     \\
\midrule
              & \multicolumn{7}{c}{Medium}                                                                                                                                                                \\
\midrule
GPEN \cite{yang2021gan}          & 25.1871                  & 0.7460                   & 0.3063                    & 0.6741                  & 12.2091                 & 5.3168                 & 3.6872                     \\
GFPGAN \cite{wang2021towards}        & 26.2826                  & 0.7839                   & 0.2554                    & 0.6970                  & \blue{11.1332}                 & \blue{5.2539}                 & 3.7173                     \\
RestoreFormer \cite{wang2022restoreformer} & 25.5123                  & 0.7256                   & 0.3346                    & \blue{0.7044}                  & 11.2567                 & 5.3760                 & \blue{3.6448}                     \\
CodeFormer \cite{zhou2022codeformer}    & 24.6238                  & 0.7424                   & 0.2852                    & 0.6077                  & 11.8149                 & 6.7256                 & 3.7899                     \\
EDVR \cite{wang2019edvr}          & 26.3385                  & 0.7815                   & 0.2625                    & 0.6771                  & 12.4233                 & 5.2598                 & 3.6660                     \\
BasicVSR \cite{chan2021basicvsr}      & 25.8332                  & 0.7725                   & 0.2594                    & 0.6638                  & 12.4503                 & 5.7990                 & 3.8101                     \\
BasicVSR++ \cite{chan2022basicvsr++}   & \blue{26.5465}                  & \blue{0.7918}                   & \blue{0.2203}                    & 0.6919                  & 13.4386                 & 6.8957                 & 5.6914                     \\
\textbf{KEEP (Ours)}          & \red{27.4853}                  & \red{0.8171}                   & \red{0.1740}                    & \red{0.7481}                  & \red{9.5937}                  & \red{4.6179}                 & \red{3.3764}                     \\
\midrule
              & \multicolumn{7}{c}{Heavy}                                                                                                                                                                 \\
\midrule
GPEN \cite{yang2021gan}          & 25.0191                  & 0.7437                   & 0.3108                    & 0.6544                  & 12.4814                 & 5.6768                 & 3.8088                     \\
GFPGAN \cite{wang2021towards}        & 26.0747                  & 0.7807                   & 0.2613                    & \blue{0.6761}                  & \blue{11.6804}                 & 6.8689                 & 3.9346                     \\
RestoreFormer \cite{wang2022restoreformer} & 25.3354                  & 0.7216                   & 0.3458                    & 0.6715                  & 11.7674                 & \blue{5.6277}                 & \blue{3.6966}                     \\
CodeFormer \cite{zhou2022codeformer}    & 24.5600                  & 0.7407                   & 0.2916                    & 0.5949                  & 12.0462                 & 5.9110                 & 3.9079                     \\
EDVR \cite{wang2019edvr}          & 26.1600                  & 0.7792                   & 0.2729                    & 0.6524                  & 13.0927                 & 5.9243                 & 3.8166                     \\
BasicVSR \cite{chan2021basicvsr}      & 25.6895                  & 0.7695                   & 0.2686                    & 0.6426                  & 12.7841                 & 6.1689                 & 3.8356                     \\
BasicVSR++ \cite{chan2022basicvsr++}    & \blue{26.2686}                  & \blue{0.7872}                   & \blue{0.2289}                    & 0.6650                  & 14.2254                 & 7.2980                 & 6.1919                     \\
\textbf{KEEP (Ours)}          & \red{27.2165}                  & \red{0.8124}                   & \red{0.1803}                    & \red{0.7282}                  & \red{9.8833}                  & \red{4.8643}                 & \red{3.3217}                     \\
\bottomrule
\end{tabular}
% \vspace{-0.4cm}
\end{table*}
Table \ref{tab:full_sota} provides the full quantitative results of models on different test partitions. As can be observed, our proposed method consistently outperforms all other concurrent approaches on all datasets.
In particular, KEEP surpasses BasicVSR++ \cite{chan2022basicvsr++} by a large margin of $0.95$ dB in PSNR on test dataset with heavy degradation.
For identity preservation and pose quality metrics (\textit{IDS} and \textit{AKD}), our method achieves top performance and fewer fluctuations. On VFHQ-mild dataset, KEEP possesses $8.82$ average keypoint distances on images of shape $512\times 512$, while the distances of all other methods are over $10.53$.
This suggests that our method could better preserve identity within the generated video and introduce far less jitters in the pose of faces.
Such improvements are significant in VFSR.

\section{Limitations and Future Work}
\begin{figure}[!t]
    \centering
    \includegraphics[width=.9\linewidth]{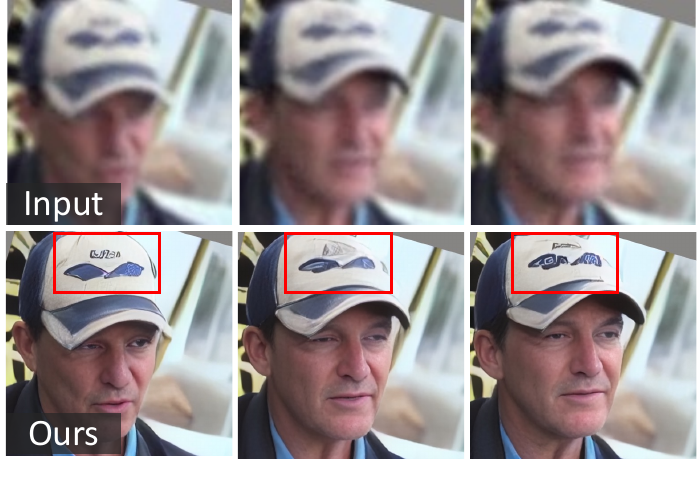}
    % \vspace{-0.2cm}
    \caption{
        \textbf{Limitations.}
        Our method might produce inconsistent results on non-facial areas when the input video exhibits heavy degradation.
        For example, the logo on the hat shows various shapes in different frames.
    }
    \label{fig:limitations}
    % \vspace{-0.5cm}
\end{figure}

Figure~\ref{fig:limitations} presents a failure case of our method when the input video suffers heavy degradation.
The recovered logo on the hat in different frames shows inconsistent shapes.
This could stem from the inherent limitation that the contents in non-facial areas are unstructured and highly deviate from what the facial prior code encapsulates. 
A potential solution is to use general well-trained VSR models to further enhance these regions and backgrounds. We leave this avenue of research as future work.

\section{Evaluation on Real-World Videos}
\begin{figure}[!t]
    \centering
    \includegraphics[width=.98\linewidth]{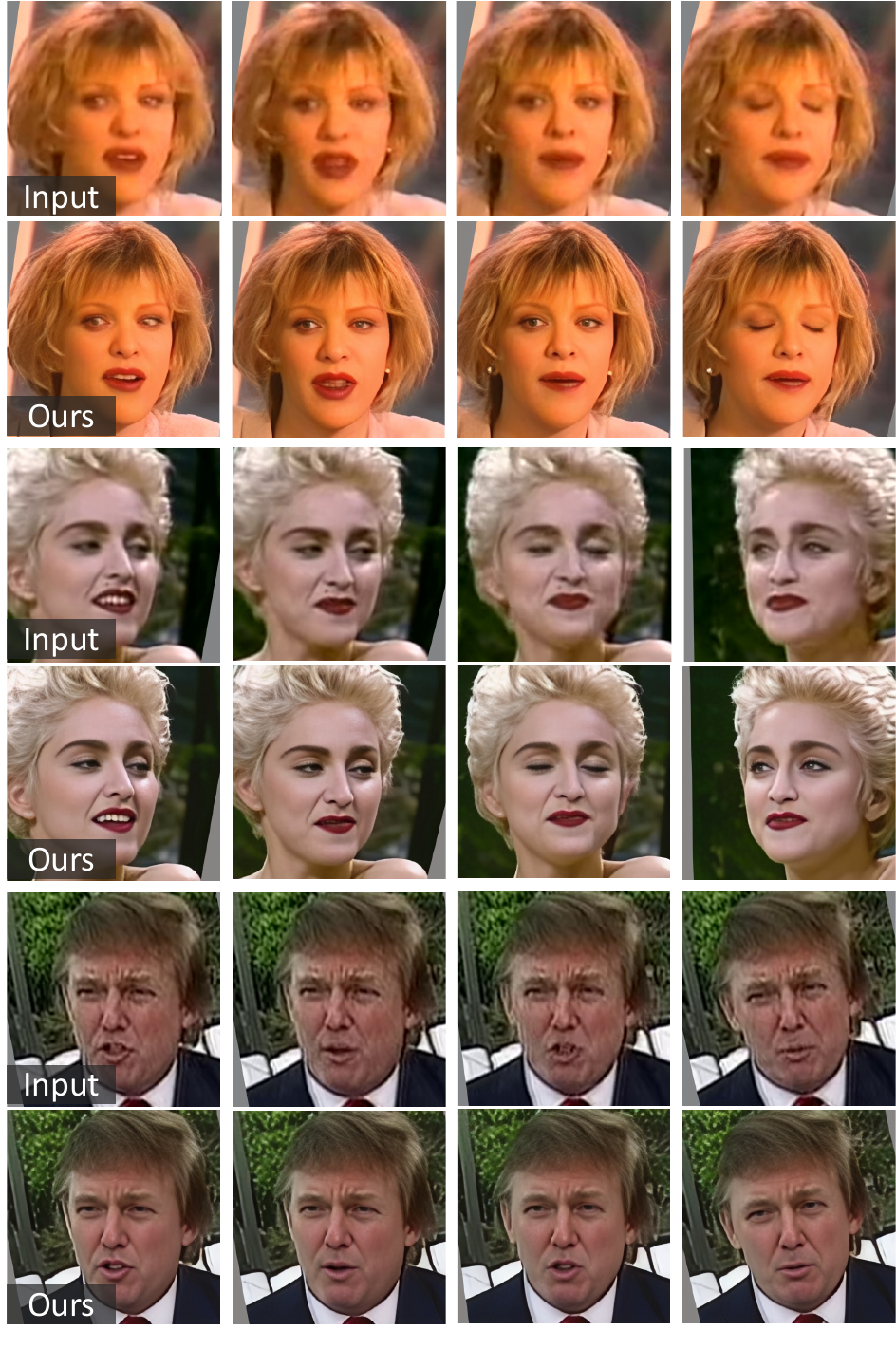}
    % \vspace{-0.2cm}
    \caption{
        \textbf{Qualitative comparison on the real face videos.}
        Our KEEP recovers high-fidelity face videos with faithful and consistent details.
    }
    \label{fig:real_vis}
    % \vspace{-0.5cm}
\end{figure}

Figure~\ref{fig:real_vis} shows that our method recovers texture details in each frame. 
In addition, the supplementary video delivers high-quality restoration of our method with superior consistency from highly degraded face videos, demonstrating extraordinary generalization in face videos in the wild.

\section{Additional Visual Results}
\begin{figure*}[t]
    \centering
    \includegraphics[width=.98\linewidth]{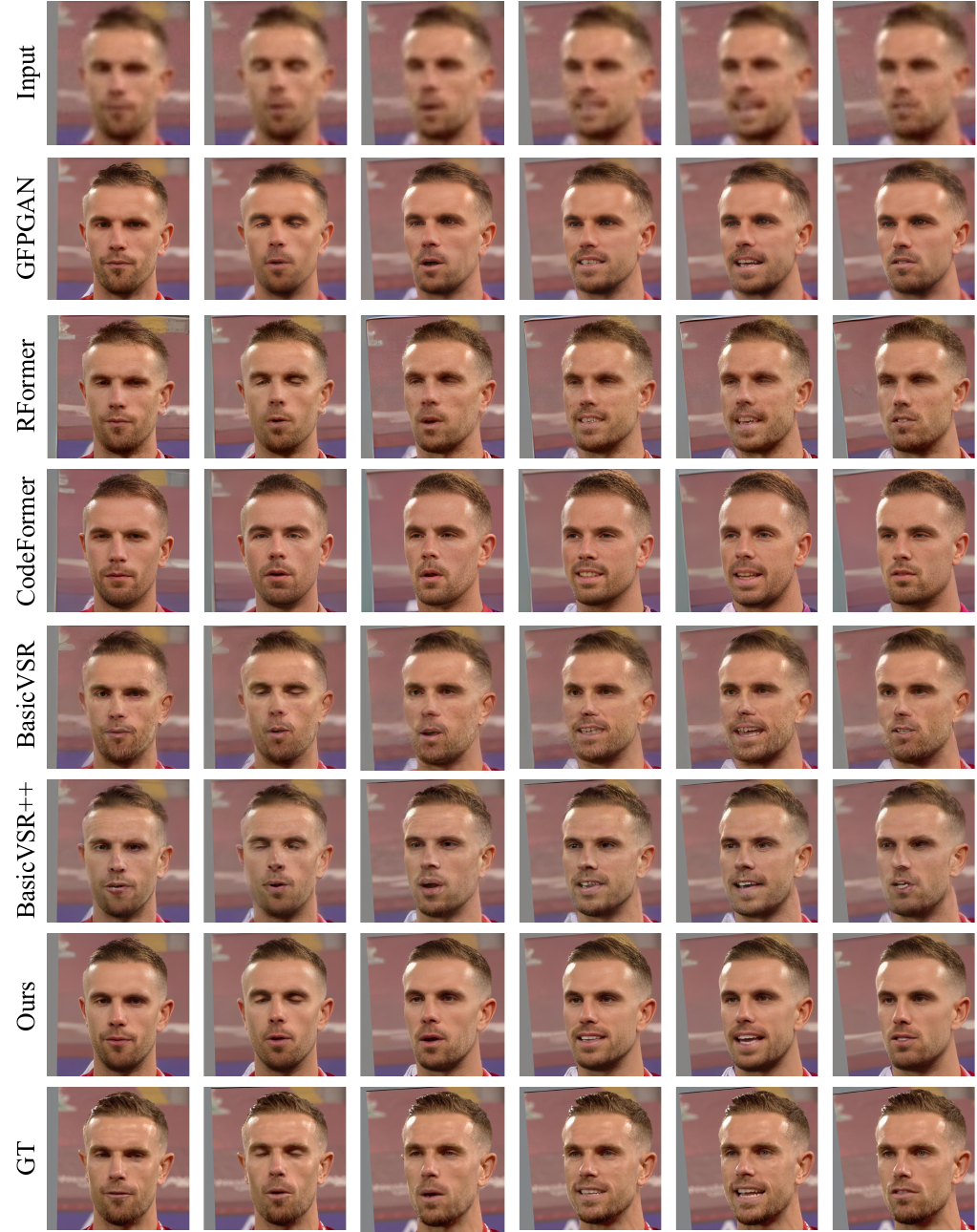}
    % \vspace{-0.2cm}
    \caption{
        \textbf{Qualitative comparison on the VFHQ.}
        RFormer represents RestoreFormer~\cite{wang2022restoreformer}.
    }
    \label{fig:sota_comp1}
    % \vspace{-0.5cm}
\end{figure*}

\begin{figure*}[t]
    \centering
    \includegraphics[width=.98\linewidth]{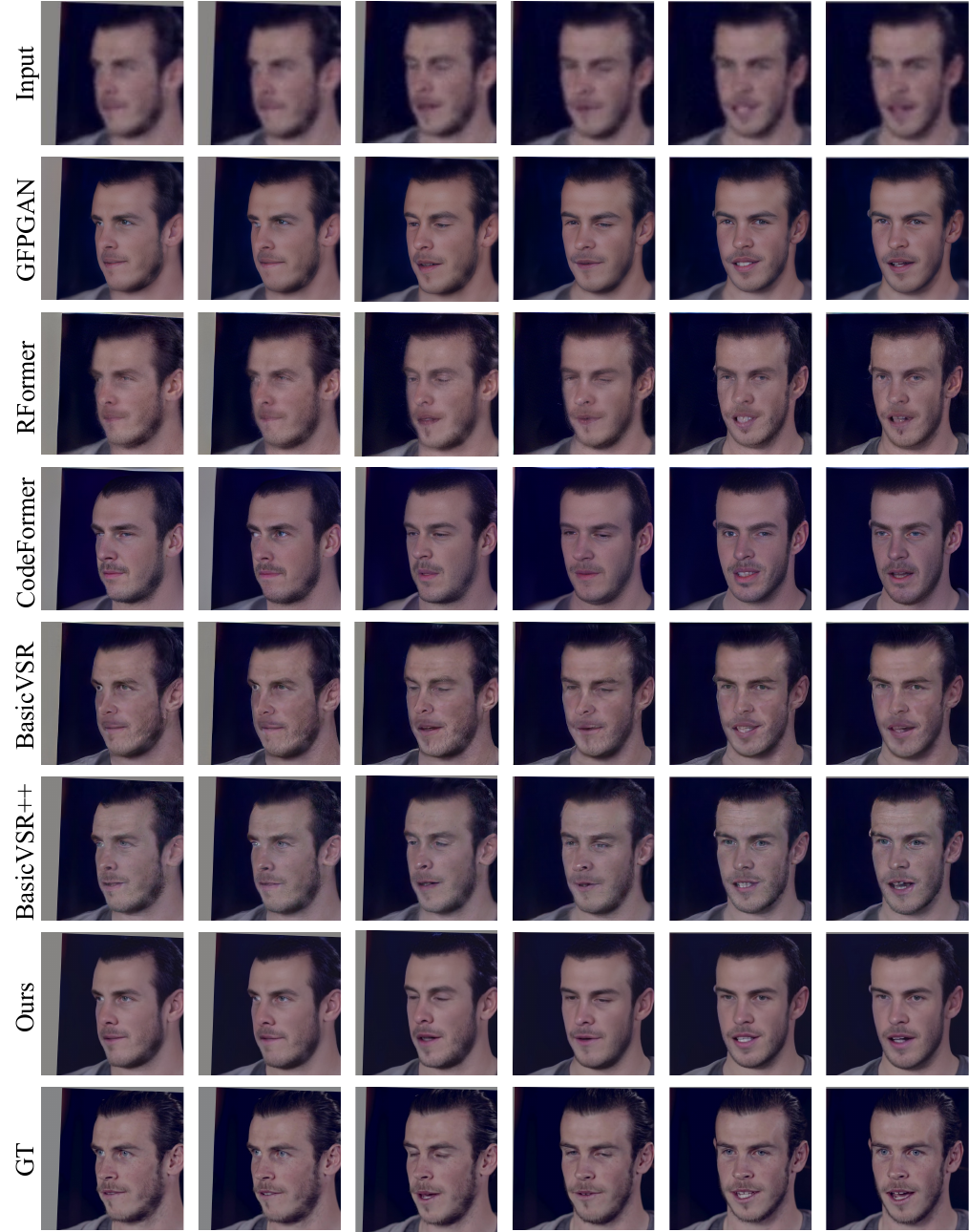}
    % \vspace{-0.2cm}
    \caption{
        \textbf{Qualitative comparison on the VFHQ.}
        RFormer represents RestoreFormer~\cite{wang2022restoreformer}.
    }
    \label{fig:sota_comp2}
    % \vspace{-0.5cm}
\end{figure*}

\begin{figure*}[t]
    \centering
    \includegraphics[width=.98\linewidth]{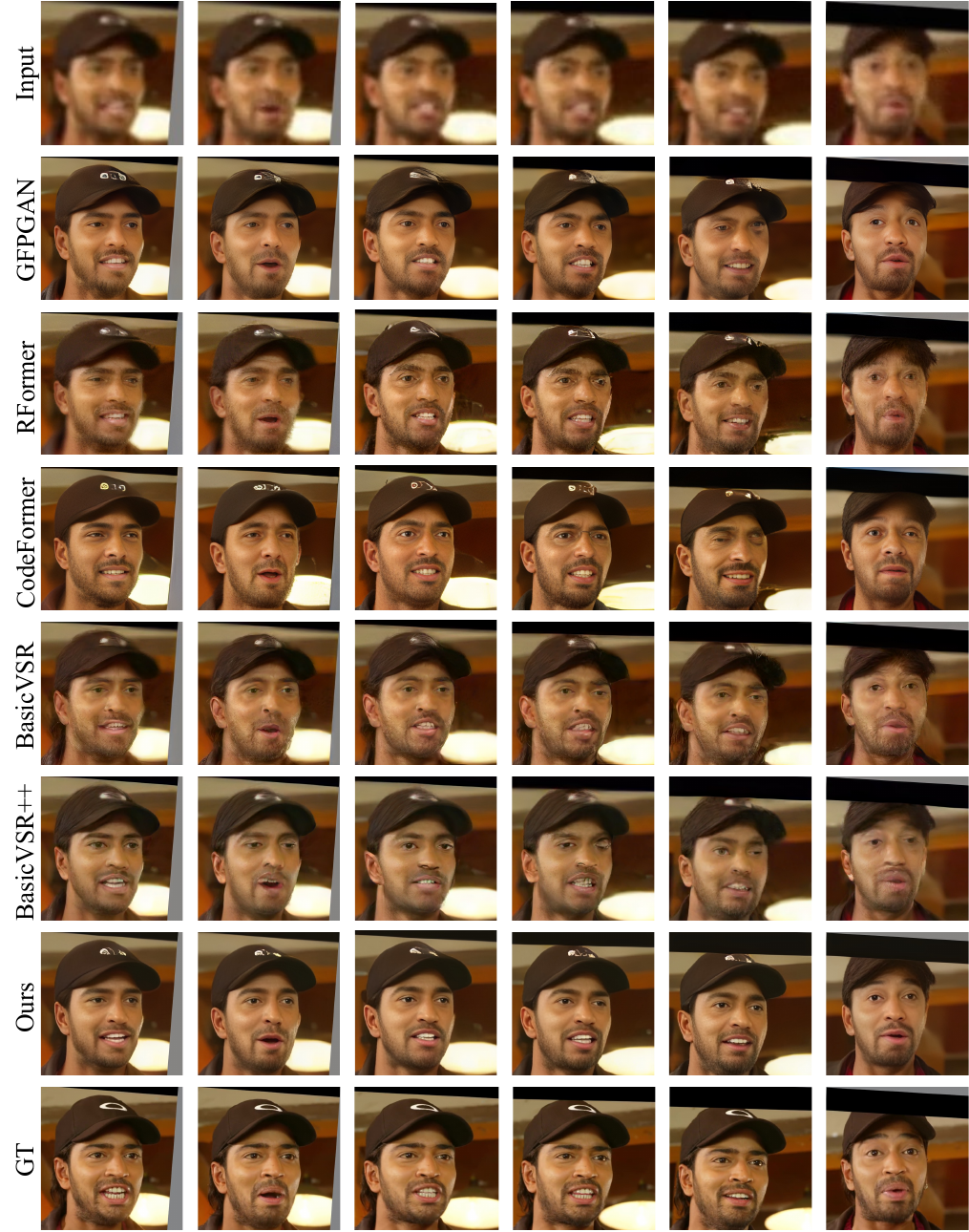}
    % \vspace{-0.2cm}
    \caption{
        \textbf{Qualitative comparison on the VFHQ.}
        RFormer represents RestoreFormer~\cite{wang2022restoreformer}.
    }
    \label{fig:sota_comp3}
    % \vspace{-0.5cm}
\end{figure*}

\begin{figure*}[t]
    \centering
    \includegraphics[width=.98\linewidth]{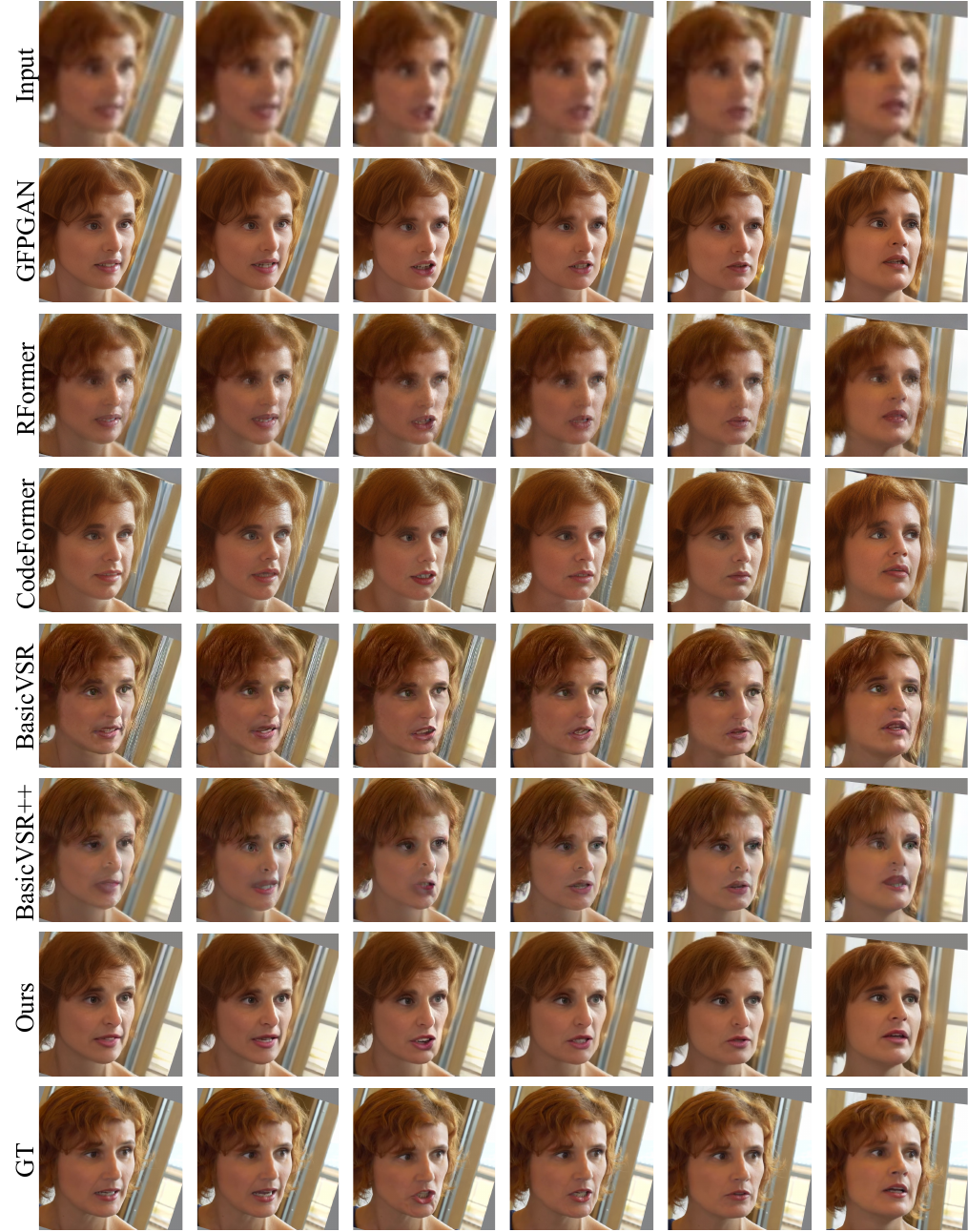}
    % \vspace{-0.2cm}
    \caption{
        \textbf{Qualitative comparison on the VFHQ.}
        RFormer represents RestoreFormer~\cite{wang2022restoreformer}.
    }
    \label{fig:sota_comp4}
    % \vspace{-0.5cm}
\end{figure*}

Figure \ref{fig:sota_comp1}, \ref{fig:sota_comp2}, \ref{fig:sota_comp3}, and \ref{fig:sota_comp4} showcase additional visual examples of our methods and other compared baselines.


% \clearpage  % TODO REVIEW/FINAL: This \clearpage needs to be removed from both review and camera-ready versions.

% ---- Bibliography ----
%
% BibTeX users should specify bibliography style 'splncs04'.
% References will then be sorted and formatted in the correct style.
%
\bibliographystyle{splncs04}
\bibliography{main}